%% file: main.tex
\documentclass[format=manuscript, screen=true, review=false, nonacm=true]{acmart}
\newcommand{\doctitle}{Communication-Efficient Clustered Federated Learning in Non-IID Wireless Networks}
\input{includes/acm_header}

\input{includes/macros_acm}

\input{includes/macros_nonacm}
\input{figures/sequence_defs}

  \usepackage[caption=false,font=footnotesize,labelfont=sf,textfont=sf]{subfig}

\usepackage[USenglish]{babel}

\GlsXtrLoadResources[
      src={includes/acronyms},
      type=acronym,
      sort={en-GB}
      ]

\GlsXtrLoadResources[
      src={includes/symbols},
      type=symbols,
      sort={en-GB}
]

 \GlsXtrLoadResources[
      src={includes/terms},
      sort={en-GB}
 ]

\begin{document}
\input{includes/acm_authors}
\begin{abstract}%
\input{chapters/abstract}\end{abstract}%

\maketitle

\glsresetall

\section{Introduction}
\input{chapters/introduction}

\section{Background}
\input{chapters/background}
\section{Methods}\label{sec:methods}
\input{chapters/methods}
\section{Experiments}
\input{chapters/datasets}

\input{chapters/results}

\section{Related work}
\input{chapters/relatedwork}

\section{Discussion and limitations}\label{sec:discussion}
\input{chapters/discussion}

\section{Conclusion and future work}
\input{chapters/conclusion}

\section*{Acknowledgment}
\addcontentsline{toc}{section}{Acknowledgment}
\input{includes/acknowledgment}

\bibliographystyle{ACM-Reference-Format}
\bibliography{IEEEabrv,references/main}%

\cleardoublepage%
\appendix
\begin{appendices}
      \input{chapters/appendix}

      \cleardoublepage%
      \printunsrtglossary[type=acronym]%
      \printunsrtglossary[type=main, title=Terminology]
      \printunsrtglossary[type=symbols, title=Symbols]
\end{appendices}
\end{document}

%% file: includes/acm_header.tex
\setcopyright{rightsretained}
\copyrightyear{2026}
\acmYear{2026}
\acmDOI{To be added}

%% file: includes/macros_acm.tex
\usepackage{bm}
\usepackage{amsmath}
\usepackage{makecell}

\usepackage[online, neutral, extended]{ericolors}
\usepackage[round-mode=places,detect-weight=true,round-pad = false,locale=US]{siunitx}  %
\DeclareSIUnit\px{px}
\DeclareSIUnit\samples{data~samples}
\DeclareSIUnit\pp{pp}

\usepackage[
    acronym,
    abbreviations,
    shortcuts,
    symbols,
    record,
    section=section
  ]{glossaries-extra}

\newglossary*{terminology}{Terminology}

\setabbreviationstyle{long-short}
\setabbreviationstyle[acronym]{long-short}

\setglossarystyle{index}

\usepackage[capitalise]{cleveref}  %

\AtBeginEnvironment{appendices}{\crefalias{section}{appendix}}

\crefname{appendix}{Appendix}{Appendices}
\Crefname{appendix}{Appendix}{Appendices}

\usepackage{mathtools}
\usepackage{algorithm}

\usepackage[inline]{enumitem}

\usepackage{multirow}

\usepackage{xspace}

\newcommand{\cifar}{\mbox{\glsxtrshort{CIFAR}-10}}
\newcommand{\ifca}{\mbox{\glsxtrshort{IFCA}}}
\newcommand{\ifcaep}{\mbox{\glsxtrshort{IFCA}${}_{\varepsilon}^{\!\star}$}}
\newcommand{\ifcae}{\mbox{\glsxtrshort{IFCA}${}_{\varepsilon}$}}
\newcommand{\ifcap}{\mbox{\glsxtrshort{IFCA}${}^{\!\star}$}}

\newcommand{\moe}{\gls{MoE}}
\newcommand{\pacfl}{\gls{PACFL}}
\newcommand{\moes}{\mbox{\gls{MoE}${}^{\star}$}}

\newcommand{\femnist}{\glsxtrshort{FEMNIST}}
\newcommand{\deepmimo}{\glsxtrshort{DeepMIMO}}
\newcommand{\rotatedcifar}{Rotated \cifar{}}
\newcommand{\ucbone}{\texttt{UCB1}}

\newcommand*{\eg}{e.g.\@\xspace}

\def\etal.{et\penalty50\ al.}
\newcommand{\di}{{\mathrm{i}\mkern1mu}}

\newcommand{\R}{\mathbb{R}}
\newcommand{\C}{\mathbb{C}}

\newcommand{\flconstfnt}[1]{\ensuremath{\mathsf{#1}}}
\newcommand{\given}{\mathrel{}\middle\vert\mathrel{}}

\DeclareMathOperator{\softmax}{\gls{softmax}}
\DeclareMathOperator*{\argmax}{arg\,max}
\DeclareMathOperator*{\argmin}{arg\,min}

\let\arccos\relax
\DeclareMathOperator{\arccos}{arc\,cos}

\makeatletter
\newcommand*{\transpose}{%
  {\mathpalette\@transpose{}}%
}
\newcommand*{\@transpose}[2]{%
  \raisebox{\depth}{$\m@th#1\intercal$}%
}

\makeatother

\newcommand{\norm}[1]{\lVert#1\rVert}

\newcommand\hlf[1]{%
    \tikz[baseline,%
      outer sep=-15pt, inner sep = 2pt%
    ]%
   \node[rounded corners,rectangle,fill=EricssonTeal,anchor=text]{#1\xspace};%
}%

\newcommand{\LComment}[1]{$\triangleright$ {#1}}
\newcommand{\runin}[1]{\textbf{#1.}\space}

%% file: includes/macros_nonacm.tex
\usepackage{microtype}
\microtypecontext{spacing=nonfrench}
\microtypesetup{
      activate={true, nocompatibility},
      protrusion=false,
}

\usepackage{appendix}
\usepackage{algpseudocode}
\usepackage{nth}

\usepackage{tikz}
\usetikzlibrary{quotes,bending}
\usetikzlibrary{patterns,shapes.arrows, arrows.meta, fit, calc, intersections}
\usetikzlibrary{shapes, decorations, arrows, angles}
\usetikzlibrary{3d, backgrounds, decorations.text}
\usetikzlibrary{positioning, shapes.symbols, tikzmark, decorations.markings}
\usetikzlibrary{decorations.pathreplacing, calligraphy, perspective}

\tikzset{weird fill/.style={text=white,
        text height=24pt,
        text width=55mm,
        minimum width=73.6mm, minimum height=20mm,
        inner xsep=24pt, append after command={
   \pgfextra
        \fill[sharp corners, fill=#1]%
    (\tikzlastnode.west)%
    [rounded corners=4pt] |- (\tikzlastnode.north)%
    [rounded corners=0pt] -| (\tikzlastnode.east)%
    [rounded corners=0pt] |- (\tikzlastnode.south)%
    [rounded corners=4pt] -| (\tikzlastnode.west);
   \endpgfextra}}}
\tikzset{mydeco/.style={pic actions/.append code=\tikzset{postaction=decorate}}}

\usepackage{pgfplots}
\usepgfplotslibrary{dateplot}

\pgfplotsset{compat=newest,
    width=10cm,
    height=5cm,
    scale only axis,
    try min ticks=7,
    every axis/.style={
        axis y line=left,
        axis x line=bottom,
        axis line style={thick,->,>=latex, shorten >=-.4cm},
        mark size=.5,
        label style={font={\tiny\sffamily}},
        tick label style={font=\tiny\sffamily}
    },
    tick style={black, thick},
    colorbar style={at={(1.25, 1)}}
}
\tikzset{
    semithick/.style={line width=0.8pt},
}

\usepackage{tikzscale}

\usepackage{adjustbox}
\usepackage{array}
\usepackage{arydshln}

\usepackage[disable, textsize=tiny]{todonotes}
\usepackage{sparklines}
\usepackage{interval}

\newcommand{\BigO}[1]{\ensuremath{\gls{bigO}\!\left(#1\right)}}

%% file: figures/sequence_defs.tex
\newlength{\textheightbox}

\normalsize %
\newsavebox{\CBox}
\sbox{\CBox}{S} %
\tikzset{%
  >={Latex[width=2mm,length=2mm]},
    base/.style = {rectangle,
        rounded corners=2.5pt,
        draw=none,
        text=black,
        minimum width=\textheightbox,
        minimum height=\textheightbox,
        font=\sffamily,
        outer sep=0,
        inner sep=0},
    server/.style = {base, text=white, fill=EricssonBlue},
    client/.style = {base, text=white, fill=EricssonTeal},
    description/.style = {
        base,
        inner sep=0,
        text width=0.7\linewidth,
        minimum width=0.7\linewidth,
        anchor=base west,
        xshift=0.05\linewidth,
        yshift=-\textheightbox,
        font=\rmfamily},
    eriline/.style = {
        draw=EricssonBlack,
        thick,
        -latex,
        rounded corners=2.5pt
    },
    serverline/.style = {
        dotted
    }
}

\def\drop{\baselineskip}        %
\def\leftclear{\textheightbox}   %

\newcommand{\serverdot}{\tikz[baseline=-0.5ex]\node[server]{};}
\newcommand{\clientdot}{\tikz[baseline=-0.5ex]\node[client]{};}

%% file: includes/acm_authors.tex
\title{\doctitle}

\author{Martin Isaksson}
\email{martin.isaksson@ericsson.com}
\orcid{0000-0001-9972-0179}
\affiliation{%
  \institution{Ericsson AB and KTH Royal Institute of Technology}
  \country{Sweden}
}

\author{Rickard~Cöster}
\orcid{0000-0001-9671-7919}

\affiliation{%
  \institution{Ericsson AB}
  \city{Stockholm}
  \country{Sweden}
}

\author{\v{S}ar\={u}nas~Girdzijauskas}
\affiliation{%
  \institution{KTH Royal Institute of Technology and RISE Research Institutes of Sweden}
  \city{Stockholm}
  \country{Sweden}
}
\orcid{0000-0003-4516-7317}

\author{Seif Haridi}
\affiliation{%
  \institution{KTH Royal Institute of Technology and RISE Research Institutes of Sweden}
  \city{Stockholm}
  \country{Sweden}
}
\orcid{0000-0002-6718-0144}

\renewcommand{\shortauthors}{Isaksson et al.}

%% file: chapters/abstract.tex
Current and emerging mobile networks face increasing privacy, scalability, and
performance challenges as distributed learning becomes essential for advanced
\glsxtrshort{6G} services. %
These challenges are amplified by the presence of non-\glsxtrshort{IID} data,
naturally arising in heterogeneous wireless environments, that
degrade the performance of standard \glsxtrfull{FL} approaches. %
To address non-\glsxtrshort{IID} data across clients, we propose a
\gls{CFL} framework that assigns clients to specialized models
during training. Data-domain clustering initializes model-based
clustering, enabling identification and exploitation of client similarities
without sharing raw data while reducing communication cost.

Motivated by \glsxtrshort{mmWave} beam selection, we evaluate our methods on a
wireless simulation dataset (\deepmimo{}) and complement it with two
non-\glsxtrshort{IID} benchmarks (\femnist{}, \rotatedcifar{}) to isolate
different non-\glsxtrshort{IID} characteristics. Our best configurations improve
accuracy over \ifca{} by \qty{4.8}{\pp} on \deepmimo{} and \qty{10.6}{\pp} on
\femnist{} while reducing \glsxtrlong{DL} data volume by \qty{83.1}{\percent} on
\femnist{} compared to \ifca{}.

We demonstrate that different clustering strategies suit different
non-\glsxtrshort{IID} characteristics. These results show that our \gls{CFL}
methods enable resource-efficient, and high-performance
distributed learning in future \glsxtrshort{6G} networks.

%% file: chapters/introduction.tex
The evolution from \nth{5} to \gls{6G} mobile networks introduces unprecedented
demands on performance, capacity, coverage, and privacy~\cite{ericsson6g}. While
\glsxtrshort{5G} has significantly advanced mobile broadband and low-latency
communications, many emerging applications---such as holographic
communication, extended reality, and the Internet of
Senses~\cite{ericsson6gsenses}---will require \glsxtrshort{6G} mobile networks
that are secure, trustworthy, and
intelligent~\cite{ericsson6gspectrum,ericsson6g,ericsson6gsecurity,
saad2020a}.

\gls{mmWave} frequency bands play an important role in meeting these
requirements by enabling extremely high data rates through wide
bandwidths~\cite{ericssonmmwave, ma2023deep}. However, \gls{mmWave}
communication also introduces new challenges, including directional beam
management, link instability, and increased signaling
overhead~\cite{ma2023deep}. %
In particular, selecting the best \gls{DL} beamforming vector (see~\cref{sec:beamforming}) requires
extensive search over candidate beams, resulting in high communication overhead
and raising privacy concerns when data must be shared across base stations.

\begin{figure}
    \centering
    \includegraphics[width=0.75\linewidth]{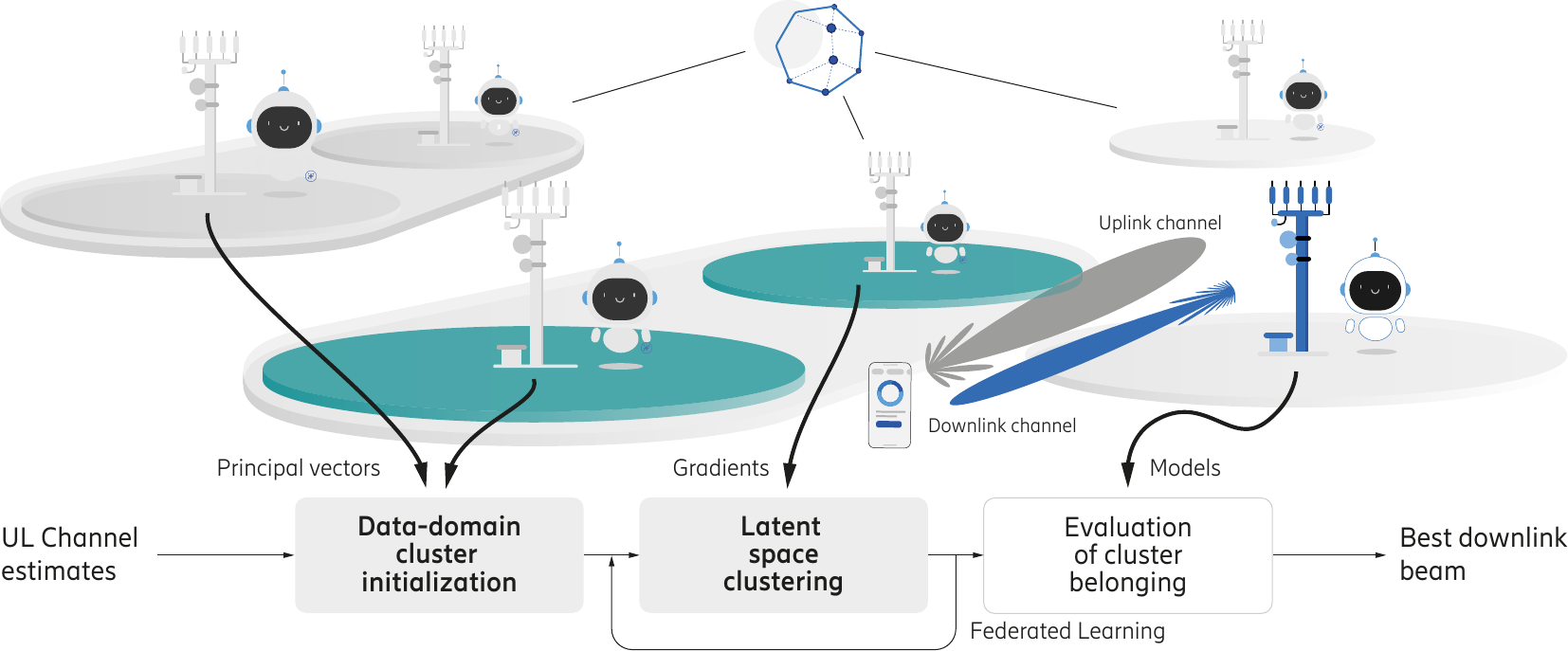}
    \caption{\textbf{Overview of the proposed training process.} In the \gls{mmWave} beam selection use-case, \gls{UL} \gls{sub6} channel estimates are acquired from sounding reference symbols and used to predict the best \gls{DL} \gls{mmWave} beam. We increase performance, robustness, and reduce overhead in non-\glsxtrshort{IID} scenarios by clustering clients (base stations) first in the data-domain~(\cref{sec:data_domain_similarity}), and second in the latent space~(\cref{sec:ifca,sec:moe}).}
    \label{fig:hero}
\end{figure}%
\glsreset{DL}%
\glsreset{UL}%
To address privacy concerns in distributed wireless systems,
\gls{FL}~\cite{mcmahan2017communication, bonawitz2019towards} has emerged
as a promising approach that enables multiple clients (\eg{}, base stations) to
collaboratively train a global model without sharing raw data. Despite its
advantages, performance of standard \gls{FL} methods such as \gls{FedAvg}
degrades significantly in non-\glsxtrshort{IID}
scenarios~\cite{zhao2018federated}, which naturally arise in heterogeneous
wireless environments due to variations in user spatial distribution, behavior,
geography, hardware, and traffic patterns. Such degradation limits model
performance and system efficiency.

\gls{CFL}~\cite{sattler2021clustered} methods such as
\gls{IFCA}~\cite{ghosh2020ifca} have emerged as promising solutions to
non-\glsxtrshort{IID} challenges. Recent
work~\cite{isaksson2023mmwave,cheng2022} demonstrated that integrating \gls{CFL}
and \gls{MoE} models can address non-\glsxtrshort{IID} challenges in
\gls{mmWave} beam selection. However, these methods suffer from high
communication overhead and computational complexity, limiting their scalability
in large-scale \gls{6G} mobile networks.

The key insight is that non-\glsxtrshort{IID} data distributions cause clients
to form clusters where the data distributions are approximately
\glsxtrshort{IID}. Determining which clients should train together is the hard
part.

\textbf{In this work,} we propose an adaptive \gls{CFL} framework that
assigns clients to cluster models, overcomes limitations on high communication
overhead and computational complexity, while also improving accuracy,
see~\cref{fig:hero}.

Our framework combines data-domain cluster initialization and model-based
clustering to capture heterogeneity across clients. The data-domain cluster
initialization provides a cheap initialization to the model-based clustering
\gls{CFL} methods that refine the assignment during \gls{FL} training.

Our methods include improved variants of \ifca{} which evaluate all global
cluster models per communication round to assign a client to a cluster, and
\gls{MoE} methods that use a local gating model in each client to assign clients
to a cluster---balancing exploration and exploitation---\emph{without}
evaluating all global cluster models in every communication round.

We evaluate our framework on \deepmimo{} for a motivational \gls{mmWave}
\glsxtrlong{DL} beam selection use-case using \glsxtrlong{UL} \gls{sub6} channel
estimates, as well as two datasets (\femnist{}, \rotatedcifar{}) that exhibit
the key non-\glsxtrshort{IID} characteristics defined
in~\cref{sec:problem_formulation}, demonstrating that our methods also apply
beyond wireless.

The proposed framework (with \moe{}) achieves \qty{10.6}{\pp} accuracy
improvement on \femnist{} compared to \ifca{}, and
\qty{83.1}{\percent} lower total \gls{DL} data volume than \ifca{}. On
\deepmimo{}, our \ifca{}-based methods improve accuracy by \qty{4.8}{\pp}
with run-to-run accuracy spread reduced up to seven-fold. To select the best
method for each dataset, we first rank methods by accuracy, and choose the
method with the lowest total \gls{DL} data volume among the methods whose
difference in accuracy is not statistically significant from the
best~(see~\cref{tbl:summary}).

\textbf{Our main contributions are:}%
\begin{enumerate}
     \item \textbf{Characteristic-driven evaluation.} We evaluate our framework
    on three datasets, \deepmimo{}, \rotatedcifar{}, and \femnist{}, chosen to
    cover different non-\glsxtrshort{IID} characteristics, demonstrating that
    the best clustering strategy depends on the dominant non-\glsxtrshort{IID}
    characteristics and that our methods generalize beyond wireless (\cref{sec:datasets,tbl:summary}).
    \item \textbf{Communication-efficient \gls{MoE} for \glsxtrshort{CFL}.} We
    introduce a \gls{MoE} framework that by selecting a \emph{single} global
    cluster model reduces \gls{DL} data volume from \(\BigO{\gls{J}}\) to
    \(\BigO{1}\) in the number of clusters~\gls{J}. This lowers total \gls{DL}
    data volume by \qty{83.1}{\percent} compared to \ifca{},
    while increasing accuracy by \qty{10.6}{\pp}~(\femnist, \cref{sec:moe,tbl:summary});
    \item \textbf{Loss-proportional cluster selection.} We introduce variants of
    \gls{IFCA} that replace argmin-loss cluster assignment with cluster
    assignment probability inversely proportional to loss, initialized from
    cheap data-domain clustering. These enhance accuracy by up to \qty{4.8}{\pp}
    compared to \ifca{}~(\deepmimo, with \gls{PACFL} initialization) and reduce the run-to-run accuracy spread
    from \num{7.37} to as little as \num{1.06} for
    \ifcap{} without \gls{PACFL} on \deepmimo{}~(\cref{tbl:deepmimo}).
\end{enumerate}

%% file: chapters/background.tex
\subsection{Analog beamforming and beam management}\label{sec:beamforming}

In \glsxtrshort{5G} and upcoming \gls{6G} networks, the use of \gls{mmWave}
frequency bands allows high throughput and low latency but introduces challenges
such as increased directionality and sensitivity to blockage. Adding antenna
elements increases the effective antenna gain and narrows beams, improving
signal strength, but requiring precise beam management. Furthermore, since more
beams are needed for coverage, beam management becomes more
time-consuming~\cite{ma2023deep,giordani2019a}.

\emph{Beamforming} is the technique to steer these narrow beams in a specific
direction. This is done by adjusting the phase and amplitude of the transmitted
signal at each antenna element in the antenna array. In \emph{analog
beamforming}, which is the focus of this work, a single \gls{RF} chain drives
all antenna elements while the phase of the signal to each antenna element is
adjusted individually by phase shifters to steer the signal in the desired
direction.

\begin{figure}
    \centering
    \adjustbox{width=.5\linewidth}{%
        \input{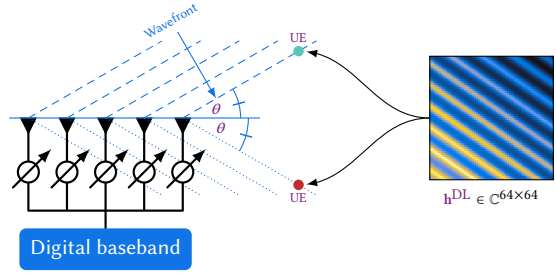}
    }
    \caption{\textbf{Illustration of the front-back ambiguity of a \gls{ULA} antenna.} Two \glspl{UE} (\textcolor{EricssonTeal}{in front of}, \textcolor{EricssonRed}{behind} the antenna)
    at the same distance from the antenna produce identical \gls{DL} channel responses~\gls{hdli}, making the \glspl{UE} indistinguishable from the channel response alone. This ambiguity creates the \emph{concept shift on labels} we observe across base stations in the \deepmimo{} simulation.}
    \label{fig:ula}
\end{figure}%

Recent
studies~\cite{alrabeiah2019deep,alrabeiah2020deep,DBLP:conf/wcnc/ChafaaNBD21,cheng2022}
have explored data-driven beam selection, leveraging \gls{UL} channel
information at \gls{sub6} frequencies to infer the best \gls{mmWave} \gls{DL}
beam. While promising, most existing works assume \glsxtrshort{IID} data across
clients, which is unrealistic in real deployments. In practice, user and
environmental conditions differ across cells, resulting in non-\glsxtrshort{IID}
data distributions that degrade model convergence and accuracy.

Each \emph{client} in the beam selection task is a base station equipped
with an antenna array~(\cref{fig:ula,fig:map}), each \emph{data sample} is a
\gls{UL} channel estimate, and each \emph{class label} is a beam index
from the \gls{mmWave} codebook~\cref{eq:codebook}.

In this work, following~\cite{isaksson2023mmwave}, we consider a more realistic
scenario with data having interesting non-\glsxtrshort{IID} characteristics such
as label distribution skew (beam index) and concept shift (same features,
different label), and we assign each data sample (\gls{UL} channel estimate) to
the strongest base station, see~\cref{fig:map}.

Specifically, for a one-dimensional \gls{ULA} antenna as seen in~\cref{fig:ula}, let a predefined quantized beam steering \gls{mmWave}
codebook~\cite{alrabeiah2019deep} be represented as~\({\gls{F}
= \left[ \gls{f}(0), \ldots, \gls{f}(\gls{beamformingindex}),\ldots,\gls{f}(\gls{nobeams}-1) \right] }\) with
\(\gls{nobeams}\) beamforming vectors as columns each corresponding to a
transmission or reception direction we call beam angle \gls{beamangle}. For \gls{transmitantennas} antenna elements spaced half a wavelength (\(\gls{lambda}/2\)) apart, a beamforming vector in~\gls{F} can be written~as%
\begin{equation}
    \gls{F}_{\gls{m},\gls{beamformingindex}} = \frac{1}{\sqrt{\gls{transmitantennas}}} e^{-\di \pi\,\gls{m}\,\cos\left.\,\gls{beamangle}\left(\gls{beamformingindex}\right)\,\right.},%
    \label{eq:codebook}
\end{equation}%
where \({\gls{m} \in \left\{0,\dots, \gls{transmitantennas}{-}1\right\}}\) is the precoder
weight index, for \gls{transmitantennas} transmit antennas.
The beam angle for beamforming vector \({\gls{beamformingindex} \in
\left\{0,\dots, \gls{nobeams}{-}1\right\}}\) is
\(\gls{beamangle}(\gls{beamformingindex})= \frac{\pi
\gls{beamformingindex}}{\gls{nobeams}}\).

\subsection{Problem formulation}\label{sec:problem_formulation}

We consider a geographically distributed and privacy-sensitive \gls{ML} scenario
with \gls{K} clients (\eg{}, base stations) \(\gls{S} = \gls{Sset}\). Each
client \(\gls{k} \in \gls{S}\) holds a local dataset~\gls{Pk} containing
\({\gls{nk} =\vert\gls{Pk}\vert}\) data samples \({\left(\gls{xi},
\gls{yi}\right)}\) which remain local and are never shared due to privacy or
business constraints.

We formulate beam selection as a multi-class classification problem where the
goal is to learn a global model \(\gls{wg} \in \R^ {\gls{d}}\) that minimizes
the expected population loss,\nopagebreak%
\begin{equation}
\min_{\gls{wg} \in \R^{\gls{d}}} \gls{flf}(\gls{wg}) \text{, where} \, \gls{flf}(\gls{wg}) \overset{\Delta}{=} \underbracket[.12pt][5pt]{\sum_{\gls{k}=1}^{\gls{K}} \frac{\gls{nk}}{\gls{n}} \overbracket[.12pt][6pt]{\frac{1}{\gls{nk}} \sum_{\gls{i} \in \gls{Pk}} \underbracket[.12pt][6pt]{\gls{l}\left(\gls{xi}, \gls{yi}, \gls{wg}\right)}_{\text{sample \gls{i} loss}}}^{\text{client \gls{k} average loss}}}_{\text{population average loss}},
\label{eq:loss}
\end{equation}
where \({\gls{l}\left(\gls{xi}, \gls{yi}, \gls{wg}\right)}\) is the
negative log-likelihood loss function for the ground truth~\({\gls{yi}}\) and the predicted \({\gls{yhatg} = \gls{fg}\left(\gls{xi}, \gls{wg}\right)}\).
In other words, we aim to minimize the average loss of the global model
over all clients in the population.

The first goal is to achieve high performance on the local test set of each
client under non-\glsxtrshort{IID}
settings~\cite{DBLP:conf/icml/HsiehPMG20,kairouz2021advances,Huyen22,breitholtz2025adaption,elrifai2025survey,briggs2020federated},
such~as:\nopagebreak%
\begin{itemize}
\item \emph{Feature distribution skew} (covariate-shift). The feature
distributions are different between clients,
so that the marginal distributions \(\gls{P}\left(\gls{x}\right)\) vary,
but \(\gls{P}\left(\gls{y} \given \gls{x}\right)\) is shared;
\item \emph{Label distribution skew} (prior probability shift, or class imbalance).
    The distribution of class labels varies among clients, so
    that \(\gls{P}\left(\gls{y}\right)\) varies but \(\gls{P}\left(\gls{x} \given \gls{y} \right)\)
    is shared;
\item \emph{Same label, different features} (concept shift on features). The conditional distributions
\(\gls{P}\left(\gls{x} \given \gls{y}\right)\) varies between clients but
\(\gls{P}\left(\gls{y}\right)\) is shared;
\item \emph{Same features, different labels} (concept shift on labels). The conditional distribution \(\gls{P}\left(\gls{y} \given \gls{x}\right)\)
varies between clients, but \(\gls{P}\left(\gls{x}\right)\) is shared;
\item \emph{Quantity skew} (unbalancedness). Clients have different amounts of data.
\end{itemize}

To solve this under non-\glsxtrshort{IID} settings we apply \gls{CFL} that
learns \gls{J} global cluster models and a client-to-cluster assignment~\({\gls{c}:\gls{S} \to \gls{Jremset}}\),
\begin{equation}
    \min_{\{\gls{wgj} \in \R^{\gls{d}}\},\,\gls{c}}\ \sum_{\gls{k}=1}^{\gls{K}}
\frac{\gls{nk}}{\gls{n}}\, \frac{1}{\gls{nk}}\sum_{\gls{i}\in\gls{Pk}}
\gls{l}\left(\gls{xi},\gls{yi},\gls{wgck}\right),
\end{equation}
which reduces to~\eqref{eq:loss} for \(\gls{J}=1\).

The second goal is to reduce communication cost, specifically the total \gls{DL}
data volume, i.e.\ the number of models downloaded times the model size, summed
over clients and communication rounds.

Our datasets (see~\cref{sec:datasets}) and methods (see~\cref{sec:methods}) span
these key non-\glsxtrshort{IID} challenges: \deepmimo{} introduces primarily
concept shift on labels stemming from the beam selection use-case,
see~\cref{fig:map}; \rotatedcifar{} exhibits exhibits feature distribution skew
and concept shift on features due to the rotation of the input data from a
subset of the clients; and \femnist{} demonstrates multiple characteristics
stemming from the handwriting of individuals including concept shift on
features, feature distribution skew (covariate shift), label distribution skew,
and quantity skew. Together, they provide a comprehensive benchmark without
implying exhaustive coverage of all possible non-\glsxtrshort{IID} scenarios.

\subsection{\glsfmtfull{FL}}\label{sec:fl}

\gls{FL}~\cite{mcmahan2017communication} enables collaborative model training across clients without sharing raw data. A central parameter server coordinates training on many clients over several communication rounds
until convergence.

In \gls{FedAvg}~\cite{mcmahan2017communication}, training of a global
model \(\gls{fg}(\gls{x}, \gls{wg})\) is distributed and
synchronous.

In each communication round \gls{t}:\nopagebreak\vspace{\baselineskip}

\noindent\input{figures/fedavg_steps.tex}

This baseline algorithm, known as \gls{FedAvg}, works efficiently in
\glsxtrshort{IID} settings but performs poorly when local datasets differ significantly
and show non-\glsxtrshort{IID} characteristics~\cite{zhao2018federated},
see~\cref{sec:problem_formulation}. In such cases, local updates diverge,
leading to slow convergence and reduced model accuracy.

To mitigate these effects, advanced approaches introduce personalization and
clustering, grouping clients with similar data or model characteristics.
However, \gls{CFL} methods may introduce new costs: loss-based clustering
frameworks such as~\ifca{}~\cite{ghosh2020ifca} require clients to download and
evaluate multiple global cluster models in each communication round; server-side
methods~\cite{sattler2021clustered} require server-side access to gradients,
which makes them incompatible with standard privacy mechanisms such as~\gls{SA}.
Both are core challenges for \gls{FL} in \gls{6G}~\cite{liu2020federated} that
this paper aims to overcome.

%% file: chapters/methods.tex
\subsection{Overview}

\gls{CFL} methods enable personalization based on heterogeneous client data
distributions. We propose a modular \gls{CFL} framework designed to operate
efficiently under non-\glsxtrshort{IID} data. The framework consists of three
components:

\begin{enumerate}
  \item Data-domain clustering for initialization (\cref{sec:data_domain_similarity});
  \item Adaptive model-based clustering during federated training (\cref{sec:ifca});
  \item Communication-efficient \gls{MoE} (\cref{sec:moe}).
\end{enumerate}

These components are modular and can be used either independently or jointly.
See pseudocode provided in~\cref{sec:appmethods}.

We aim to find clusters, or subsets of clients, that benefit from
training together---for example in terms of increased accuracy and reduced
\gls{DL} data volume---as opposed to training a single global model. Note that
clients may belong to different clusters at different times during the training
process as their models or data distributions evolve.

In general, clustering can be done in the latent space using model parameters or
gradients, using training loss, or in the data-domain on input
data~\cite{elrifai2025survey}. In some domains, clustering can be done using
available metadata such as geographical distance~\cite{iyer2018mitigating} or
categorical variables such as software versions and device
identifiers~\cite{reisser2021moe}. The telecom domain is rich in such
side-information---for example configuration parameters and performance
indicators.

As an example of clustering in the data-domain, in
\gls{PACFL}~\cite{vahidian2023efficient} the authors perform clustering in the
data-domain using a low-rank approximation of client data via truncated
\gls{SVD}. Data-domain clustering
provides informed initial cluster assignments---at the risk of locking in
clients into suboptimal clusters before model-based signals emerge. To isolate
the value of data-domain cluster initialization, we compare four initialization strategies seen in~\cref{tbl:overviewdatamethods}.

\begin{table}[tbh]
\centering
\caption{\textbf{Overview of data-domain cluster initialization methods.}}\label{tbl:overviewdatamethods}
\begin{tabular}{cm{0.5\textwidth}}
  \toprule
  \textbf{Method} & \textbf{Description} \\
  \midrule
  None & A single predefined cluster.\\
  Random & Initial client cluster assignment is random. \\
  Oracle & Initial clusters are defined by a human expert.\\
  \gls{PACFL}~\cite{morafah2022rethinking, vahidian2023efficient} & Data-domain clustering using principal angle similarity between client data subspaces~(\cref{sec:data_domain_similarity}).\\
  \bottomrule
\end{tabular}
\end{table}

\gls{IFCA}~\cite{ghosh2020ifca} (see~\cref{sec:ifca}) requires
clients to download and evaluate all global cluster models in every
communication round.

Unlike prior work~\cite{ghosh2020ifca, isaksson2023adaptive,isaksson2023mmwave}
that relies solely on model updates, we combine data-domain cluster
initialization and model-based clustering into a modular \gls{CFL} framework,
see~\cref{fig:hero}. Cluster initialization reduces \gls{DL} data volume, mainly
by re-evaluating all cluster models less often, while maintaining or increasing
accuracy, see~\cref{sec:results}. See~\cref{tbl:overviewmodelmethods} for an
overview of model exploration methods.

\begin{table}[tbh]
\centering
\caption{\textbf{Overview of model clustering methods} (without data-domain initialization). We modify \ifca{} in two ways, this table illustrates the differences. \gls{rconv} denotes the number of communication rounds to convergence and \gls{J} the number of global cluster models. Values of \gls{eps}, \gls{epssim}, and \gls{moeeps} per method are listed in~\cref{sec:hyperparams}}\label{tbl:overviewmodelmethods}
\begin{tabular}{ccccc}
  \toprule
  \textbf{Method}
      & \makecell{\textbf{$\varepsilon$-greedy}\\\textbf{exploration}}
      & \makecell{\textbf{Model-based}\\\textbf{Exploration}}
      & \makecell{\textbf{Selection}\\\textbf{rule}}
      & \makecell{\textbf{\gls{DL} downloads}\\\textbf{(total)}}\\
  \midrule
  \makecell{Data-domain\\initialization,\\such as\\\pacfl{}~\cite{morafah2022rethinking, vahidian2023efficient}} & &  & data-domain             & \gls{rconv} \\
  \ifca{}~\cite{ghosh2020ifca}          & 0         & 1             & argmin-loss             & \gls{rconv}\gls{J}\\
  \ifcae{}~\cite{isaksson2023adaptive}  & \gls{eps} & \gls{epssim}  & argmin-loss             & $\gls{rconv}\left[1+\left(1-\gls{eps}\right)\gls{epssim}\left(\gls{J}-1\right)\right]$\\
  \ifcap{} (ours)                       & 0         & 1             & $\propto 1/\text{loss}$ & \gls{rconv}\gls{J}\\
  \ifcaep{} (ours)                      & \gls{eps} & \gls{epssim}  & $\propto 1/\text{loss}$ & $\gls{rconv}\left[1+\left(1-\gls{eps}\right)\gls{epssim}\left(\gls{J}-1\right)\right]$\\
  \moe{} (ours)                         & \gls{moeeps} &            & gate softmax            & \gls{rconv}\\
  \moes{} (ours)                        & \gls{moeeps} &            & gate + \glsxtrshort{UCB}       & \gls{rconv}\\
  \bottomrule
\end{tabular}
\end{table}

\subsection{Data-domain cluster initialization}\label{sec:data_domain_similarity}
Our aim is to find a similarity measure between two clients that captures
the statistical heterogeneity between different client data domains. In
principle, this can be any available side-information~\cite{reisser2021moe}.

\subsubsection{Local subspace representation}

\begin{figure}
  \centering
  \begin{adjustbox}{width=.35\linewidth}%
    \includegraphics{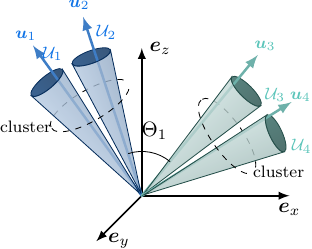}%
  \end{adjustbox}%
  \caption{\textbf{The smallest principal angle between client data subspaces}
  is used to capture similarity between clients. $\gls{leftsingvec}_{\gls{k}}$
  are examples of principal vectors in the client subspaces
  \gls{clientsubspacek}. The smallest principal angle
  \({\gls{smallestprincipalangle}\left(\gls{clientsubspacek},
  \gls{clientsubspacel} \right)}\) is the smallest angle between any pair of
  principal vectors from the two subspaces spanned by clients \gls{k} and
  \gls{kp}. It defines the similarity between these clients and is used by the
  server to cluster clients into disjoint clusters.}
  \label{fig:subspaces}
\end{figure}

In this work we define a similarity measure in the data subspaces. Inspired by
\gls{PACFL}~\cite{morafah2022rethinking, vahidian2023efficient}, we leverage the
angle between data subspaces, see~\cref{fig:subspaces}, spanned by the
\gls{principalcomponents} most significant left singular vectors of a local data
matrix of each client. The local data matrix aggregates all client training set
samples, one per column. In the case of \deepmimo{}, a data sample is an
\gls{UL} channel estimate, which means that the subspace captures the dominant
channel directions for each client. Base stations with similar dominant channel
directions are expected to be clustered together, so that our similarity measure
effectively groups base stations with similar local channel geometry. In the
\deepmimo{} Outdoor~\num{1} Blockage scenario, dominant channel directions
correlate with geographic \gls{UE} distributions, so base stations with similar
local geographic \gls{UE} distributions cluster together even when they serve
different areas of the map.

To get the most significant vectors, we take the truncated~\gls{SVD} of the
local data matrix \(\gls{Mk} \in \R^{\gls{nofeatures} \times \gls{nk}}\) (or in
the case of \deepmimo{} \(\gls{Mk} \in \C^{\gls{nofeatures} \times
\gls{nk}}\)), where \gls{nofeatures} is the number of features. The truncated~\gls{SVD} of a matrix \gls{Mk} is a
factorization of the form%
\begin{equation}
\gls{Mtilde} = \gls{U}_{\gls{k},\gls{principalcomponents}}\gls{singularvaluematrix}_{\gls{k},\gls{principalcomponents}}\gls{rightsingularvaluematrix}_{\gls{k},\gls{principalcomponents}}^\transpose,
\end{equation}
where \(\gls{U}_{\gls{k},\gls{principalcomponents}} \in \R^{\gls{nofeatures}
\times \gls{principalcomponents}}\) (or
\(\gls{U}_{\gls{k},\gls{principalcomponents}} \in \C^{\gls{nofeatures} \times
\gls{principalcomponents}}\) using conjugate transpose for complex-valued data)
is an orthonormal matrix containing the \gls{principalcomponents} dominant left
singular vectors. Its columns span the client subspace \gls{clientsubspacek}. We
set the number of principal vectors \gls{principalcomponents} to retain, on
average, \qty{80}{\percent} (a common threshold in dimensionality reduction) of
the explained variance of the local data, for example
\({\gls{principalcomponents}=56}\) for \deepmimo{}, to balance information
preservation with communication efficiency. Sharing \gls{principalcomponents}
principal vectors reduces communication overhead compared to sharing raw data.
However, sharing principal vectors still leak information about client data
distribution~\cite{kim2023exploring}. Formal privacy analysis is orthogonal to
our contributions and can be addressed by additional mechanisms such as
differential privacy and \gls{SA}~\cite{kairouz2021advances}.

\subsubsection{Principal-angle similarity}

Let \gls{clientsubspacek} and \gls{clientsubspacel} be the subspaces spanned by
clients \gls{k} and \gls{kp}~\cite{morafah2022rethinking}. The smallest
principal angle between them is%
\begin{equation}
\gls{smallestprincipalangle}\left(\gls{clientsubspacek},  \gls{clientsubspacel} \right) = \min_{\gls{leftsingvec} \in \gls{clientsubspacek}, \gls{leftsingvec}' \in \gls{clientsubspacel}} \arccos \left(\frac{\left\vert\left\langle \gls{leftsingvec} \given \gls{leftsingvec}' \right\rangle\right\vert}{\norm{\gls{leftsingvec}}\norm{\gls{leftsingvec}'}}\right),
\end{equation}
where the minimizing pair \(\gls{leftsingvec}_1\), \(\gls{leftsingvec}'_1\) are the corresponding principal vectors. We now construct an adjacency matrix \gls{adjacencymatrix} where%
\begin{equation}
\gls{adjacencymatrix}_{\gls{k},\gls{kp}}=\gls{smallestprincipalangle} \left(\gls{clientsubspacek},  \gls{clientsubspacel} \right),\; \gls{k},\gls{kp} \in \{1, 2, \ldots, \gls{K}\}.
\end{equation}
In our motivational wireless application~(\cref{sec:beamforming}), the columns
of the local data matrix \gls{M} are complex \gls{UL} channel estimates
\gls{hul}. Performing a truncated~\gls{SVD} of \gls{M} and taking its \gls{principalcomponents} left-most singular vectors then gives the
dominant spatial directions (beams) of these channels. Each singular value then
corresponds to how much of the channel power lies along that direction.

\begin{equation}
      \gls{hul} \approx \gls{U}_{\gls{principalcomponents}}\gls{singularvaluematrix}_{\gls{principalcomponents}}\gls{rightsingularvaluematrix}_{\gls{principalcomponents}}^\dagger,
\end{equation}

where \(\gls{U}_{\gls{principalcomponents}} \in \C^{m \times
\gls{principalcomponents}}\) and
\(\gls{rightsingularvaluematrix}_{\gls{principalcomponents}} \in
\C^{n \times \gls{principalcomponents}}\) have orthonormal columns,
\(\gls{singularvaluematrix}_{\gls{principalcomponents}}\) is a
\(\gls{principalcomponents} \times \gls{principalcomponents}\)
diagonal matrix with non-negative real numbers (singular values) on the main
diagonal~\cite{tse2005fundamentals}. The principal-angle similarity has therefore a physical meaning. Base stations
whose \glspl{UE} are served from similar directions span similar subspaces.

\subsubsection{Clustering of similarity matrices}

We linearly transform the angle matrix \gls{adjacencymatrix} into a
similarity matrix \(1-
\frac{\gls{adjacencymatrix}}{\max{\left(\gls{adjacencymatrix}\right)}}\) and use
that as a precomputed affinity matrix as input to a spectral clustering
algorithm~\cite{shi2000normalized,luxburg2007tutorial}. This inverts angles (smaller angles meaning  higher similarity) and normalizes to
\(\interval{0}{1}\), where \num{1} indicates identical client subspaces and
\num{0} indicates the most dissimilar pair of clients observed.

The number of clusters \gls{J} is determined using the eigengap heuristic (the
largest difference between consecutive eigenvalues) on the similarity matrix,
which reveals natural groupings in the data~\cite{luxburg2007tutorial}.

We use spectral clustering to construct an \emph{initial estimate} of clusters
and assign clients accordingly. This is in contrast to
\gls{PACFL}~\cite{vahidian2023efficient} which uses hierarchical clustering. The
spectral clustering algorithm performs a low-dimensional embedding of the
similarity matrix, followed by clustering (here \gls{KMeans}) of the components
of the eigenvectors in the low-dimensional space. In the rest of the paper,
\gls{PACFL} refers to this variant: principal-angle similarity followed by
spectral clustering. We provide a more detailed description
in~\cref{alg:fl.preclustering,alg:fl.truncsvd}.

In summary, we perform data-domain similarity and clustering before \gls{CFL} training:\par
\nopagebreak
\vspace{\baselineskip}
\noindent\input{figures/pacfl_steps.tex}

\subsection{Adaptive model-based clustering}\label{sec:ifca}
\subsubsection{Standard \glsxtrshort{IFCA}}

Using \gls{IFCA}~\cite{ghosh2020ifca} we set the expected largest number of
clusters to be \gls{J} and initialize one global cluster model with weights \gls{wgj}
per cluster \({j \in \left\{1,2,\ldots,\gls{J}\right\}}\).

In each communication round \gls{t}:\par
\nopagebreak
\vspace{\baselineskip}
\noindent\input{figures/ifca_steps.tex}%

In Step 2, each selected client~\gls{k} locally performs a cluster assignment,
where it selects the global cluster model \(\gls{jhat}^k\) that has the lowest
estimated loss on the local training set.

\subsubsection{Modified \ifca{} variants}

We make two changes compared to~\cite{ghosh2020ifca}.

\runin{\ifcae} First, we add the \gls{epsgreedy} method
from~\cite{isaksson2023adaptive} so that with probability~\gls{eps} we select a
global cluster model randomly. This has the effect of reducing local computation
and \gls{DL} data transfer since we only need to download one global cluster
model---we will call this method~\ifcae{}. With probability
\((1-\gls{eps})\gls{epssim}\) we download all \gls{J} global cluster models and
select the one with the lowest estimated loss on the local training set, indexed
\gls{jhat}. Otherwise we return the previously assigned global cluster model (a
small tweak from~\cite{isaksson2023adaptive}). If no global cluster model has
been assigned we use the global model assigned by the data-domain cluster
initialization. This reduces the risk of the mode collapse of standard \ifca{}.

\runin{\ifcap}\label{sec:ifcap} Second, we select a global cluster model \gls{jhat} with a probability inversely proportional to the loss
of each global cluster model, evaluated locally such that
\begin{align}
\gls{bl} & \leftarrow \left[\left(\sum_{i \in \gls{Pk}} \gls{l}\left(\gls{xi}, \gls{yi}, \gls{wgj}\right)\right)^{-1}\right]_{j \in \gls{Jremset}}\\
\gls{jhat} & \text{ sampled from } \gls{Jremset}\,\text{with probability}\, \propto \gls{bl}.
\label{eq:rho}
\end{align}

With probability \(1-\gls{epssim}\), the previously assigned cluster model for this client is returned instead.

\runin{\ifcaep} Finally, we combine these methods into a new modified \ifca{}
variant that we call \ifcaep{}. We provide a summary of our methods
in~\cref{tbl:overviewmodelmethods}.

In Step 4, the cluster model parameters \gls{wgj} at time \(t+1\) are updated using only updates from clients that selected the \gls{j}th  cluster model, so that (using
model averaging~\cite{mcmahan2017communication,ghosh2020ifca})%
\begin{align}
\gls{nj} \leftarrow & \sum_{\gls{k} \in \left\{\gls{kp} \in \gls{St} \given \gls{jhat}^{\gls{kp}} = j\right\}} \gls{nk},\\
\gls{wgjt1} \leftarrow & \sum_{\gls{k} \in \left\{\gls{kp} \in \gls{St} \given \gls{jhat}^{\gls{kp}} = j\right\}} \frac{\gls{nk}}{\gls{nj}}\gls{wkt1}.
\end{align}

In each communication round \gls{t}:\par
\nopagebreak
\vspace{\baselineskip}
\noindent\input{figures/ifcaep_steps.tex}%

\subsection{Communication-efficient \glsxtrfull{MoE}}\label{sec:moe}

\subsubsection{Local gating model}

To address the \gls{IFCA} issue of the overhead of downloading all global
cluster models in each communication round, we turn our attention to
\gls{MoE}~\cite{jacobs1991adaptive}. \gls{MoE} is an \gls{ML} method where
multiple experts are used to divide a problem space into regions. Here our
\gls{J} global cluster models take the place of expert models, and we add a
locally deployed gating model~\gls{gatek}. See~\cref{fig:moe_overview} for an
example use-case motivated by our \deepmimo{} dataset.

\begin{figure}
  \centering
  \includegraphics[width=0.65\linewidth]{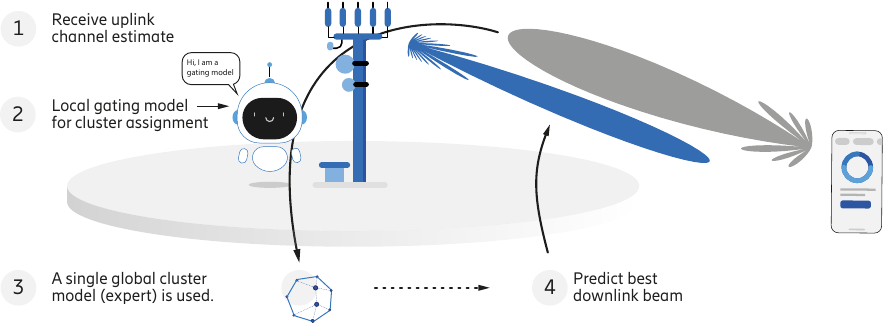}
  \caption{\textbf{Example \gls{MoE} for the \deepmimo{} use-case.} A gating model is deployed on the base station and used for cluster assignment.}
  \label{fig:moe_overview}
\end{figure}

\subsubsection{Expert selection}

In standard \gls{MoE}, a gating model weighs the different experts and mixes
their outputs. Here we use the expert weights from a \emph{local} gating model
to select a \emph{single} expert (global cluster model). We do this by selecting
the expert with the maximum weight, without needing to evaluate the experts.
This allows us to download a single expert model, reducing the per communication
round \gls{DL} data volume to that of standard \gls{FedAvg}. The idea is that
clients with similar data distributions will select the same expert in order to
avoid the degradation caused by the non-\glsxtrshort{IID} data characteristics.
The local gating models are never communicated and will remain private.

\subsubsection{Exploration and exploitation}

Because only the selected expert is evaluated, each gating model learns from
partial feedback by observing the loss of the selected expert alone. This frames
the problem as a bandit problem where each expert is an arm that receives
context \gls{x}. Without exploration, the gating model will exploit the expert
it currently believes is the best. We solve this in two ways: \moe{} and
\moes{}.

\runin{\moe{}} We use the expert weights of the local gating models as a proxy
to confidence. As the local gating models are randomly initialized, they are
initially not confident, and we use this to explore all expert models.

Each selected client evaluates its gating model, \(\gls{gateprob}(\gls{x}) =
\softmax\left(\gls{gatek}(\gls{x})\right)\), and selects an expert \gls{jhat}
based on the majority vote of the per-sample choices \(\argmax_{\gls{j}}
\gls{gateprob}_{\gls{j}}(\gls{x})\). With a probability of $\max\left(1 -
\max_{\gls{j} \in \gls{Jremset}} \bar{\gls{gateprob}}_{\gls{j}},
\gls{moeeps}\right)$, where \(\bar{\gls{gateprob}}\) is the gating output
averaged over the same samples, we instead select a random expert \({\gls{jhat}
\in \gls{Jremset}}\).

As the training progresses, the local gating models will be more and more
confident and will favor a particular expert.

\runin{\moes{}} Mode collapse and underutilization of experts are problems of
standard \gls{MoE}. In \moes{}, a bonus term is added to the output of the
gating model \({\gls{gateprob}_{\gls{j}}}\) using a \gls{UCB}
algorithm~\cite{auer2002finitetime}, specifically~\ucbone{}.

The selected expert \gls{jhat} is the expert that maximizes the sum of the
gating output and a~\ucbone{} term
\begin{align}
\gls{jhat} &= \argmax_{\gls{j} \in \gls{Jremset}} \left( \bar{\gls{gateprob}}_{\gls{j}} + \sqrt{\frac{\gls{ucbconst} \ln \gls{totalclusterusage}}{\gls{clusterusage}}} \right),
\end{align}

where \gls{clusterusage} is the number of times the \gls{j}th expert has been used by any client, and \(\gls{totalclusterusage} = \sum_{\gls{j} \in
  \gls{Jremset}} \gls{clusterusage}\). This selection rule has three major effects:
  \begin{enumerate*}[
  label=(\textit{\roman*}),
  itemjoin={{; }},
  itemjoin*={{; and }},
  after={{.}\ }
]
\item an expert is selected without needing to download any of
them%
\item the probability of selecting a specific expert depends on
the gating model confidence and how many times an expert has been
used%
\item the decision is local to the client and privacy conserving since only the
index of the selected expert is shared%
\end{enumerate*}

The exploration and exploitation trade-off is controlled by~\gls{ucbconst}. In
this work we
treat~\gls{ucbconst} as a hyperparameter (see~\cref{sec:hyperparams}). A small
value emphasizes exploitation of the best expert, and a larger value emphasizes
exploration of under-used experts. \cite{auer2002finitetime} recommend setting
$\gls{ucbconst} = 2$ when the gating outputs $\gls{gateprob}_{\gls{j}}$ are
normalized. Note here that we have set \gls{ucbconst} differently,
see~\cref{sec:hyperparams}, that experts are non-stationary and that every
expert is ensured to be selected at least once. In a real deployment we expect
hyperparameters to be tuned using network digital
twins~\cite{ohlen2022networkdigitaltwins}.

In each communication round \gls{t}:\par
\nopagebreak
\vspace{\baselineskip}
\noindent\input{figures/moe_steps.tex}

By setting~\({\gls{J}=1}\) we get \gls{FedAvg}. We add an entropy regularization
term to the loss function that sharpens the gating distribution weighted by
\gls{moealpha}, see~\cref{sec:appendix_moe} for a more
complete description.

\subsection{Baseline and experimental setup}\label{sec:training}

One way of limiting the complexity and communication overhead is to use a model
that has as few parameters as possible. For these datasets, a relatively small
\gls{CNN} architecture based on~\cite{caldas2018leaf},
see~\cref{fig:deepmimomodel} is used in conjunction with a server-side
optimization method, \gls{FedLion}, simultaneously discovered
by~\cite{isaksson2023mmwave, tang2024fedlion} to further improve performance.
Models are implemented in PyTorch~\cite{Ansel_PyTorch_2_Faster_2024} and
hyperparameters are tuned individually per dataset using~\cite{liaw2018tune}.
See~\cref{sec:hyperparams} for a list of hyperparameters.

We compare our methods against the following baselines, all using the
\gls{FedLion} optimizer~\cite{isaksson2023mmwave,tang2024fedlion} with shared
hyperparameters (except for method specific parameters,
see\cref{sec:hyperparams}) to isolate clustering effects for fair comparison:

\begin{itemize}
\item \textbf{\gls{FedAvg}}: Single global model (\gls{J}=1), no clustering
\item \textbf{\ifca{}}: Model-based clustering without data-domain initialization
\item \textbf{Random}: Random cluster assignment as lower bound
\item \textbf{Oracle}: Initial clusters are defined by a human expert (\deepmimo{} and \rotatedcifar{})
\end{itemize}

Our ablation studies
systematically evaluate: \begin{enumerate*}[
  label=(\textit{\roman*}),
  itemjoin={{; }},
  itemjoin*={{; and }},
  after={{.}\ }
]
  \item data-domain clustering alone (\gls{PACFL})
  \item model-based clustering alone (\ifca{} variants, \moe{})
  \item combined data-domain cluster initialization and model-based clustering (\gls{PACFL} + \ifca{} variants, \moe{}).
\end{enumerate*} This design allows us to attribute
performance differences to clustering architecture rather than confounding
factors.

%% file: chapters/datasets.tex
\subsection{Datasets}\label{sec:datasets}

Geographically distributed systems commonly exhibit non-identical data
distributions~\cite{DBLP:conf/icml/HsiehPMG20,kairouz2021advances,isaksson2023adaptive}
across clients due to variations in hardware, environment, and user behavior.
However, publicly available telecom datasets with sufficient scale and diversity
for evaluating distributed learning algorithms remain scarce.

To evaluate our framework under realistic non-\glsxtrshort{IID} conditions, we use
three datasets summarized in~\cref{tbl:datasets}:
\begin{enumerate*}[
  label=(\textit{\roman*}),
  itemjoin={{; }},
  itemjoin*={{; and }},
  after={{.}\ }
]
    \item \deepmimo{} (telecom
simulation)
    \item \femnist{} (handwriting)
    \item \rotatedcifar{} (image
classification)
\end{enumerate*}

\begin{figure*}
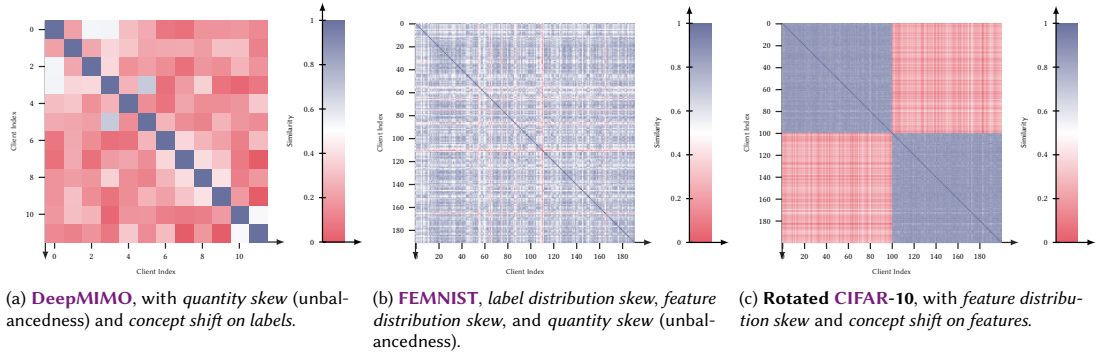

\centering

\subfloat[\textbf{\deepmimo}, with \emph{quantity
skew} (unbalancedness) and \emph{concept shift on labels}.]{\begin{adjustbox}{width=.3\linewidth}%
    \input{data/figures/heatmaps/results_deepmimo_251130_K56_J3_C06_similarity_heatmap.tikz}
\end{adjustbox}}%
\hspace{1em}%
\subfloat[\textbf{\femnist}, \emph{label distribution skew}, \emph{feature distribution skew}, and \emph{quantity skew} (unbalancedness).]{\begin{adjustbox}{width=.3\linewidth}%
    \input{data/figures/heatmaps/results_femnist_251130_J12_K4_similarity_heatmap.tikz}
\end{adjustbox}}%
\hspace{1em}%
\subfloat[\textbf{\rotatedcifar}, with \emph{feature distribution skew} and \emph{concept shift on features}.]{\begin{adjustbox}{width=.3\linewidth}%
    \input{data/figures/heatmaps/results_cifar10rot_topology_251130_195_2_similarity_heatmap.tikz}
\end{adjustbox}}%
\caption{\textbf{Visualization of similarity between clients} for our datasets using the method in~\cref{sec:data_domain_similarity}. As this method uses the features as input, the non-\glsxtrshort{IID} characteristics (concept shift on labels) of \deepmimo{} are not properly captured in this metric.}
\label{fig:heatmaps}
\end{figure*}

While \femnist{} and \rotatedcifar{} are not telecom-specific, they exhibit the
key non-\glsxtrshort{IID} characteristics defined in~\cref{sec:problem_formulation} that
we hypothesize will occur in real wireless deployments. These datasets let us
isolate and test specific non-\glsxtrshort{IID} types independently. Real-world
validation is essential to confirm whether our methods transfer to operational
wireless systems. \Cref{fig:heatmaps} shows the resulting client similarity matrices for all three datasets.

\begin{table}[tbh]
    \centering
    \caption{\textbf{Dataset summary statistics} for the number of samples per client, the number of classes, the number of clients \gls{K}, fraction of clients selected \gls{C}.}\label{tbl:datasets}
    \centering
        \input{data/tables/dataset_summary.tex}
\end{table}

\subsubsection*{\deepmimo{}}\label{sec:deepmimo}

\begin{figure}
    \centering
    \subfloat[Base station 2]{\includegraphics[width=0.25\linewidth]{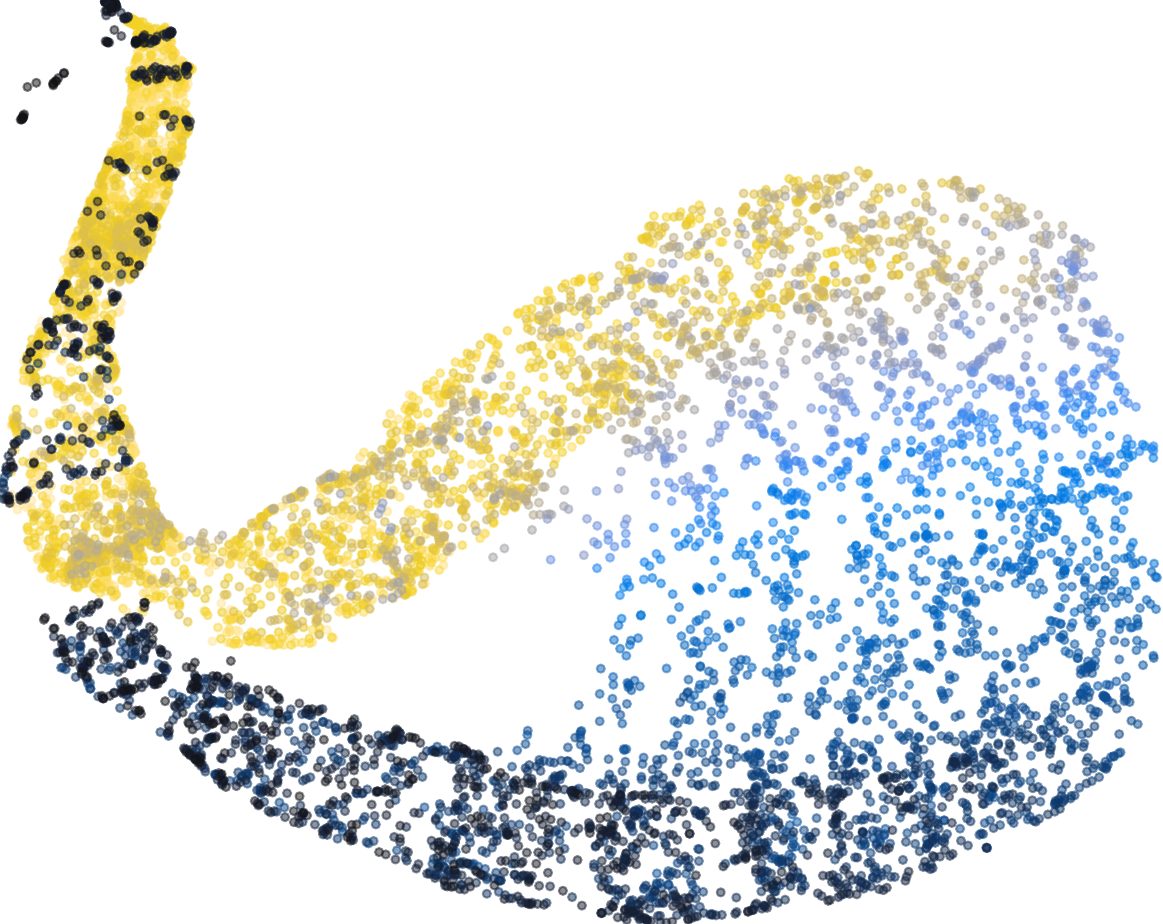}}%
    \hspace{1em}%
    \subfloat[Base station
    3]{\includegraphics[width=0.25\linewidth]{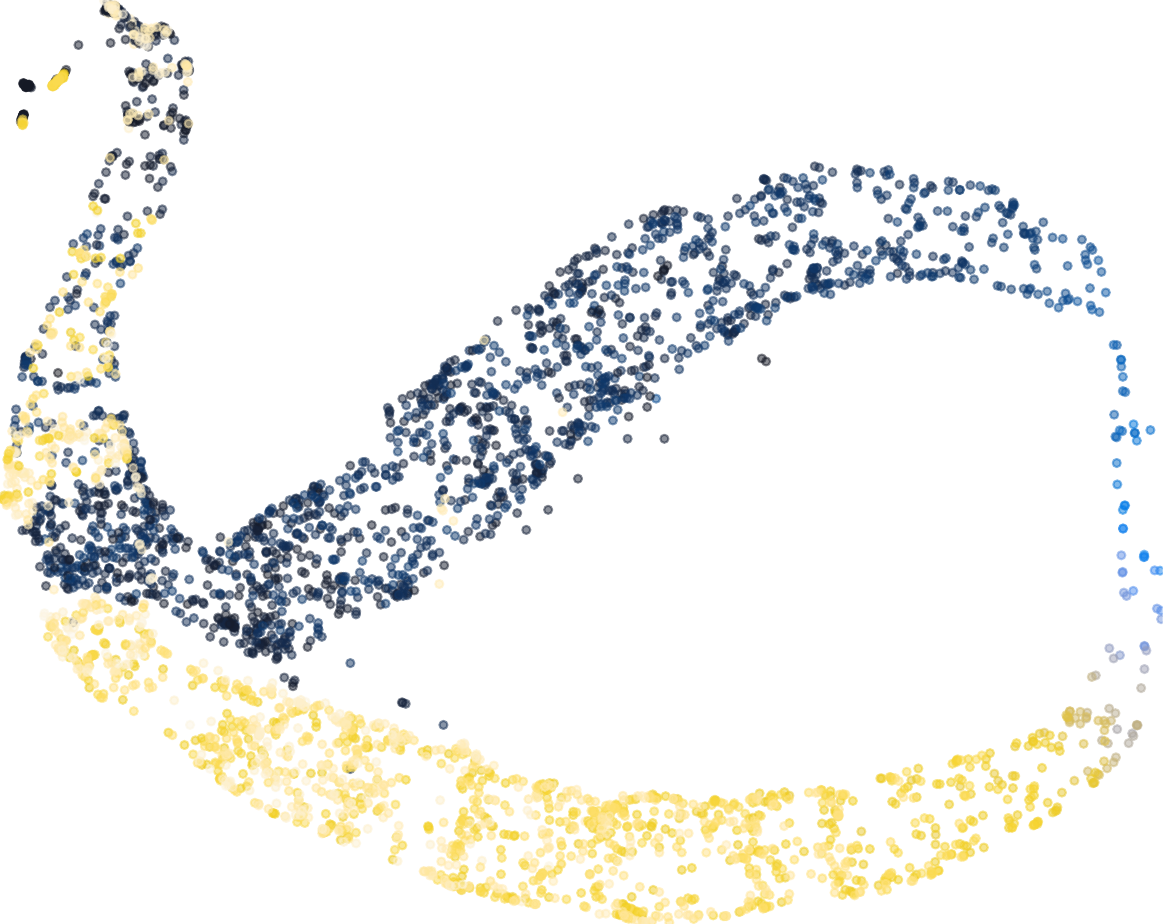}}%

    \caption{\textbf{Visualization of two neighboring base stations in the
\deepmimo{}~\cite{DBLP:journals/corr/abs-1902-06435,Remcom} dataset} using
\gls{UMAP}~\cite{mcinnes2018umap}. This dataset exhibits non-\glsxtrshort{IID}
characteristics primarily \emph{quantity skew} (unbalancedness) and
\emph{concept shift on labels} (same features, different label), here most
clearly seen in the color representing the target beamforming index. An overview
map of the dataset is shown in~\cref{fig:map}. \cref{fig:umapbs3all} shows the
embeddings for all base stations.}
    \label{fig:umapbs3}
\end{figure}

\begin{figure*}
    \centering
    \includegraphics[trim={2mm 0.45cm 2cm 0.2cm},clip, width=1\textwidth]{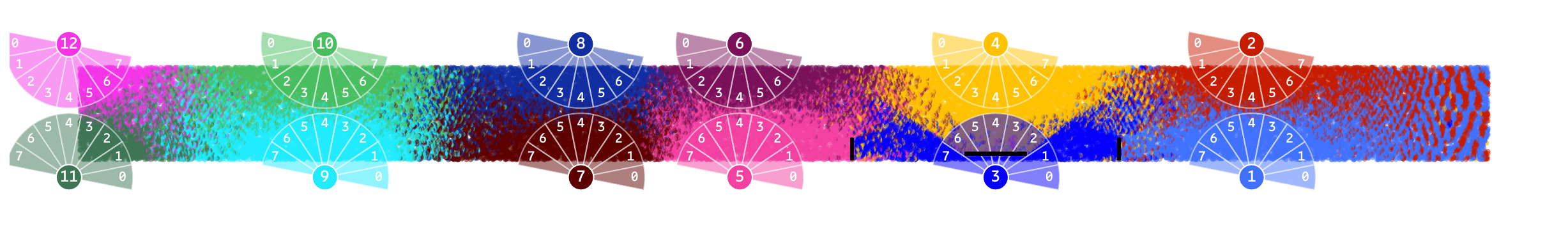}
    \caption{\textbf{The \deepmimo{}~\cite{DBLP:journals/corr/abs-1902-06435,Remcom} dataset.} We generate the \gls{mmWave} \gls{DL} channels and \gls{sub6} \gls{UL} channels for each of the 12~base stations from the Outdoor~\num{1} Blockage scenario. Each data sample is associated with the base station that has the strongest received signal in the \gls{DL}. \emph{Note the order of the beam indices,} here illustrated with \num{8} beams.}
    \label{fig:map}
\end{figure*}

\deepmimo{}~\cite{DBLP:journals/corr/abs-1902-06435} exhibits \emph{quantity
skew} (unbalancedness) and \emph{concept shift on labels} (same features,
different label), see~\cref{fig:umapbs3}, making it suitable for evaluating our
framework. The publicly available dataset generation framework \deepmimo{}
uses the Wireless InSite ray-tracing simulator~\cite{Remcom} to generate
channels between the \glspl{UE} and each base station. Specifically, we use the
Outdoor~\num{1}
Blockage scenario where we utilize base stations on either side of the main
street as clients in the \gls{FL} terminology. For each \gls{UE} position in the
map, see~\cref{fig:map}, we generate an \gls{UL} channel on a \gls{sub6}
frequency band and a \gls{DL} channel on a \gls{mmWave} frequency band. Each
\gls{UE} position is in turn associated with the base station (client) for which
the received \gls{DL} signal is the strongest without beamforming---this
results in an unbalanced dataset. For each client, we sample \qty{10}{\percent}
of the \qty{80}{\percent} of the client data set aside for training. We use
\qty{20}{\percent} of the client data as the test set for evaluation. In each
run, we sample data samples of the same size with replacement from the test set.

\begin{figure}
    \centering
    \adjustbox{width=0.75\linewidth, valign=b}{\input{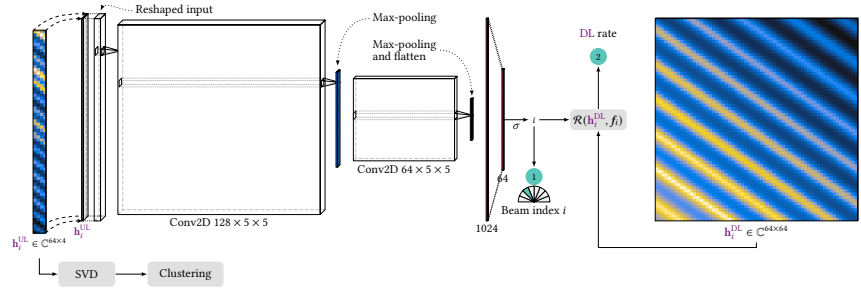}}
    \caption{\textbf{Neural Network architecture.} The complex \gls{UL} \gls{sub6} channel estimate~\gls{hul} is divided into two channels (real and imaginary) used as input to a \gls{CNN} model. The \gls{DL} \gls{mmWave} channel estimate~\gls{hdl} is only used to calculate the target beam (the label).}
    \label{fig:deepmimomodel}
\end{figure}

By taking the \gls{SVD} of the complex low-band \gls{sub6} channel matrix, we
can decompose the channel into a set of orthogonal eigenmodes. Each eigenmode is
associated with a singular value. The set of orthogonal eigenmodes represents a
set of spatial paths from the transmitter to the receiver through which signal
power can be transmitted.

An important simplification in this work is that we optimize for accuracy, which here will be a conservative measurement of performance since users may be well served in the side lobes of another beam.

\subsubsection*{\rotatedcifar{}}\label{sec:rotCifar10}

Following~\cite{ghosh2020ifca} we distribute \cifar{}~\cite{Krizhevsky2009}
among \num{200}~clients. Half of the clients receive original images and the
other half the same images rotated by \ang{180}. The client feature
distributions are controlled by rotating \cifar{} images---an example of
feature distribution skew and concept shift on features.

\subsubsection*{The \femnist{} dataset}\label{sec:femnist}

\femnist{}~\cite{caldas2018leaf,Cohen2017} is a dataset with handwritten
characters written by many writers, exhibiting many non-\glsxtrshort{IID}
characteristics, \emph{label distribution skew}, \emph{feature distribution
skew}, and \emph{quantity skew} (unbalancedness). Each input is a black and
white character, and the output is a class label.

%% file: chapters/results.tex
\subsection{Results}\label{sec:results}

\moe{} combined with \gls{PACFL} data-domain cluster initialization achieves up
to \qty{10.6}{\pp} accuracy improvement compared to \ifca{} on \femnist{}.
\ifcap{} with \gls{PACFL} improves accuracy by \qty{4.8}{\pp} on \deepmimo{}. On
\rotatedcifar{}, no method significantly improves accuracy over \ifca{}.
\gls{PACFL} alone is the best method overall, matching \ifca{} accuracy with
\qty{45.5}{\percent} lower \gls{DL} data volume. \moes{} matches accuracy, with
lower run-to-run spread.

Total \gls{DL} data volume is reduced on all three datasets. On \femnist{},
total \gls{DL} data volume is reduced by~\qty{84.1}{\percent} compared
to \ifca{} (\moes{}). \moe{} achieves a smaller reduction of
\qty{83.1}{\percent}, but with a larger accuracy gain, see~\cref{tbl:summary}.
Total \gls{DL} data volume is reduced by up to~\qty{45.5}{\percent} on
\rotatedcifar{} (\pacfl{}). \moe{} with \gls{PACFL} achieves the largest
reduction out of our methods of \qty{43.3}{\percent}, with a non-significant
accuracy decrease of \qty{1.6}{\pp} compared to \ifca{}. \deepmimo{} shows a
reduction (\qty{33.6}{\percent}) compared to \ifca{}, with the best method
\gls{PACFL}+\ifcap{},
while \moes{} (with \gls{PACFL}) achieves the largest reduction out of our
methods of \qty{47.5}{\percent}. As expected, \moe{}-based methods lose their
advantage on this dataset. Randomly selecting the cluster, which is essentially
what these methods are doing when the dominating non-\glsxtrshort{IID}
characteristic is not in the features, achieves comparable total \gls{DL} data
volume. Communication rounds and per-round \gls{DL} data volume for all datasets
are shown in~\cref{fig:rounds,fig:dataperrounddl}, and wall-clock time to
convergence and per round in~\cref{fig:conv,fig:timeperround}.

We summarize the best results per dataset in~\cref{tbl:summary} where we first
find the most accurate method for each dataset. Then we
find all methods not significantly different on accuracy (Welch's t-test,
$\gls{p} > 0.05$) from the top method and pick the method with the lowest total
\gls{DL} data volume. Full results are provided
in~\cref{tbl:deepmimo,tbl:cifar10rot,tbl:femnist} in~\cref{sec:appresults}.

\begin{table}[!h]
    \centering
    \caption{\textbf{Overview of best method for each non-\glsxtrshort{IID} type.} First we rank methods by accuracy, then find all methods not significantly different from the top method and pick the best method with the lowest total \gls{DL} data volume.}\label{tbl:summary}
    \input{data/tables/tbl_summary.tex}
\end{table}

We provide an overview of accuracy for different data and model clustering
methods in~\cref{fig:accuracy} and an overview of \gls{DL} data volume
in~\cref{fig:totaldatadl,fig:totaldatadlcdf}.

Our methods \ifcap{}, \ifcaep{} as well as  \ifcae{}~\cite{isaksson2023adaptive}
match or exceed \ifca{} accuracy across all datasets. Our methods \moe{} and
\moes{} exceed \ifca{} accuracy on \femnist{} by \qty{10.6}{\pp}
and~\qty{6.6}{\pp} respectively, while underperforming for \deepmimo{}. In this
work all experts are trained jointly with the local gating model. The
cold-started experts, combined with the concept-shift-on-labels
non-\glsxtrshort{IID} characteristics of \deepmimo{}, are a likely explanation
for the lower \gls{MoE} accuracy. Warm-starting \gls{MoE} by using expert models
pretrained using \ifca{} avoids this problem~\cite{isaksson2023mmwave} and
achieves superior performance at the cost of the reduced \gls{DL} data volume we
demonstrate. We leave the combination of warm-starting with our \gls{MoE}
formulation as future work.

\input{chapters/results_figures}

We now examine performance across the three datasets, each exhibiting distinct
non-\glsxtrshort{IID} characteristics.

\subsubsection*{Feature distribution skew} %
\femnist{} is a dataset experiencing many non-\glsxtrshort{IID} characteristics, mainly feature distribution skew (covariate shift), concept shift on features, label distribution skew, and quantity skew.
See~\cref{sec:femnist}.

A method that uses a signal in the features to perform clustering will do well
on these types of non-\glsxtrshort{IID} characteristics, as shown
in~\cref{fig:femnist_acc} where our methods \moe{}, \moes{} and \ifcae{} perform
well in terms of accuracy, followed by \gls{PACFL}. Our method \moe{} (\moe{}
with \gls{PACFL} data-domain cluster initialization) outperforms \ifca{} by
\qty{10.6}{\pp} (\qty{95}{\percent} \gls{CI} [\qty{9.13}{\pp},
\qty{12.16}{\pp}], adjusted $\gls{p} \leq 0.0001$). \moes{} (\moes{}
with \gls{PACFL} data-domain cluster initialization and \gls{UCB}) outperform \ifca{} by
\qty{6.63}{\pp} (\qty{95}{\percent} \gls{CI} [\qty{4.90}{\pp},
\qty{8.36}{\pp}], adjusted $\gls{p} \leq 0.0001$). The largest improvement for \ifca{}-based methods is achieved by \ifcae{} with
\qty{5.62}{\pp} (\qty{95}{\percent} \gls{CI} [\qty{3.96}{\pp},
\qty{7.27}{\pp}], adjusted $\gls{p} \leq 0.0001$).

We compare accuracy with and without \gls{PACFL} data-domain cluster
initialization across all methods on \femnist{}. \gls{PACFL} alone (without
model-based clustering) improves accuracy over \gls{FedAvg} by \qty{8.3}{\pp}
(\qty{95}{\percent} \gls{CI} [\qty{6.2}{\pp}, \qty{10.4}{\pp}], adjusted $\gls{p} \leq
0.0001$) see~\cref{tbl:femnist_pacfl_accuracy}. However, when combined with
other methods, \gls{PACFL} shows mixed results.

The total \gls{DL} data volume, as seen
in~\cref{fig:femnist_totaldatadl,fig:femnist_totaldatadlcdf}, is affected by the
number of communication rounds needed to converge
(\cref{fig:femnist_rounds,fig:femnist_roundscdf}) and the \gls{DL} data volume
per round (\cref{fig:femnist_dataperrounddl}). In this regard \gls{MoE}
outperforms all \ifca{}-based methods, with \ifcae{}, \ifcaep{}
outperforming the \ifca{} baseline when combined with \gls{PACFL}, due to the
probability \((1 - \gls{eps})\gls{epssim}\) of performing an \ifca{} round. \ifcap{} also outperforms the \ifca{} baseline.

We evaluate the effect on communication rounds of \gls{PACFL} data-domain
cluster initialization across all methods on \femnist{}. \gls{PACFL} data-domain
cluster initialization method changes the number of communication rounds for
\ifca{}-based methods compared to no data-domain cluster initialization by
\qtyrange{-1.6}{3.0}{\percent}. No methods show significant improvement.
For~\moe{}-based methods, the effect is a non-significant trend toward more rounds
where \moe{} increases (\qty{11.1}{\percent} more rounds, \qty{95}{\percent} \gls{CI}
[\qty{0.9}{\percent}, \qty{21.2}{\percent}], adjusted $\gls{p} \approx 0.13$), while
\moes{} increases (\qty{5.5}{\percent} more rounds, \qty{95}{\percent} \gls{CI}
[\qty{-1.8}{\percent}, \qty{12.9}{\percent}], adjusted $\gls{p} \approx 0.28$).
\gls{PACFL} shows a mixed trend across \ifca{}-based methods, and an increasing
trend for other methods, see~\cref{tbl:femnist_pacfl_rounds}.

\subsubsection*{Concept shift on features} %

The clusters in \rotatedcifar{} are primarily showing a concept shift on
features, and as a consequence feature distribution skew, where clients have different features for the same label,
see~\cref{sec:rotCifar10}. This is a strong signal, where many methods do well.
None of our methods significantly outperforms \ifca{} on accuracy
(see~\cref{tbl:cifar10rot_ifca_accuracy}).

We compare accuracy with and without \gls{PACFL} data-domain cluster
initialization across all methods on \rotatedcifar{} in
\cref{tbl:cifar10rot_pacfl_accuracy}. \gls{PACFL} alone (without model-based
clustering) shows no significant accuracy change compared to \gls{FedAvg}
(\qty{3.1}{\pp}, \qty{95}{\percent} \gls{CI} [\qty{-0.6}{\pp}, \qty{6.9}{\pp}],
adjusted $\gls{p} \approx 0.19$). When combined with our \ifca{}-based methods
(\ifcae{}, \ifcaep{}, \ifcap{}), \gls{PACFL} shows no significant effect on
accuracy (\qtyrange{-1.7}{0.1}{\pp}) with the largest decrease for \ifcaep{} of
\qty{-1.7}{\pp} (\qty{95}{\percent} \gls{CI} [\qty{-3.9}{\pp}, \qty{0.4}{\pp}],
adjusted $\gls{p} \approx 0.19$). \moe{} with \gls{PACFL}~significantly reduces
accuracy (\qty{-1.6}{\pp}, \qty{95}{\percent} \gls{CI} [\qty{-2.1}{\pp},
\qty{-1.2}{\pp}], adjusted $\gls{p} \leq 0.0001$) relative to \moe{} without initialization, while the effect on \moes{}
is non-significant.

The total \gls{DL} data volume is
affected by the number of communication rounds needed to converge
(\cref{fig:rotcifar_rounds,fig:rotcifar_roundscdf}) and the \gls{DL} data volume
per round (\cref{fig:rotcifar_dataperrounddl}).

\subsubsection*{Concept shift on labels} %

For this dataset (see~\cref{sec:deepmimo}), where we have a strong concept shift
on labels and quantity skew we see that our methods \ifcaep{}, \ifcae{} and
\ifcap{} outperform \ifca{}.

We compare accuracy with and without \gls{PACFL} data-domain cluster
initialization across all methods on \deepmimo{}
in~\cref{tbl:deepmimo_pacfl_accuracy}. \gls{PACFL} alone (without model-based
clustering) reduces accuracy from \gls{FedAvg} by \qty{0.9}{\pp}
(\qty{95}{\percent} \gls{CI} [\qty{-1.9}{\pp}, \qty{0.1}{\pp}], adjusted $\gls{p} \approx
0.14$). Our \ifca{}-based methods (\ifcae{}, \ifcaep{}, \ifcap{}) show
significant improvements on accuracy. Combined with \gls{PACFL}
we see improvements of \qtyrange{2.5}{4.8}{\pp} with the largest improvement for
\ifcap{} of \qty{4.8}{\pp} (\qty{95}{\percent} \gls{CI} [\qty{3.2}{\pp},
\qty{6.4}{\pp}], adjusted $\gls{p} \leq 0.0001$). The effect of \gls{PACFL} is
non-significant on accuracy for these model-based clustering methods. Our
\ifca{}-based methods stabilize training and reduce the run-to-run accuracy
spread from \num{7.37} for \ifca{} to as little as
\num{1.06} for \ifcap{} without \gls{PACFL} (see~\cref{tbl:deepmimo}).

\moe{},~\moes{} show a reduction in accuracy,
see~\cref{tbl:deepmimo_ifca_pacfl_accuracy}. Our \moe{}-based method
underperforms on \deepmimo{}, consistent with feature-based clustering
struggling when the main non-\glsxtrshort{IID} characteristic is a strong
concept shift on labels.

\gls{PACFL} data-domain cluster initialization shows a clear trend towards lower
total \gls{DL} data volume (\qtyrange{18.4}{58.3}{\percent}) for all our methods
(each significant), see
\cref{tbl:deepmimo_pacfl_totaldatadl,fig:deepmimo_totaldatadlcdf}.

The total \gls{DL} data volume is affected by the number of communication rounds
needed to converge (\cref{fig:deepmimo_rounds,fig:deepmimo_roundscdf}) and the
\gls{DL} data volume per round (\cref{fig:deepmimo_dataperrounddl}).

%% file: chapters/results_figures.tex
\begin{figure*}
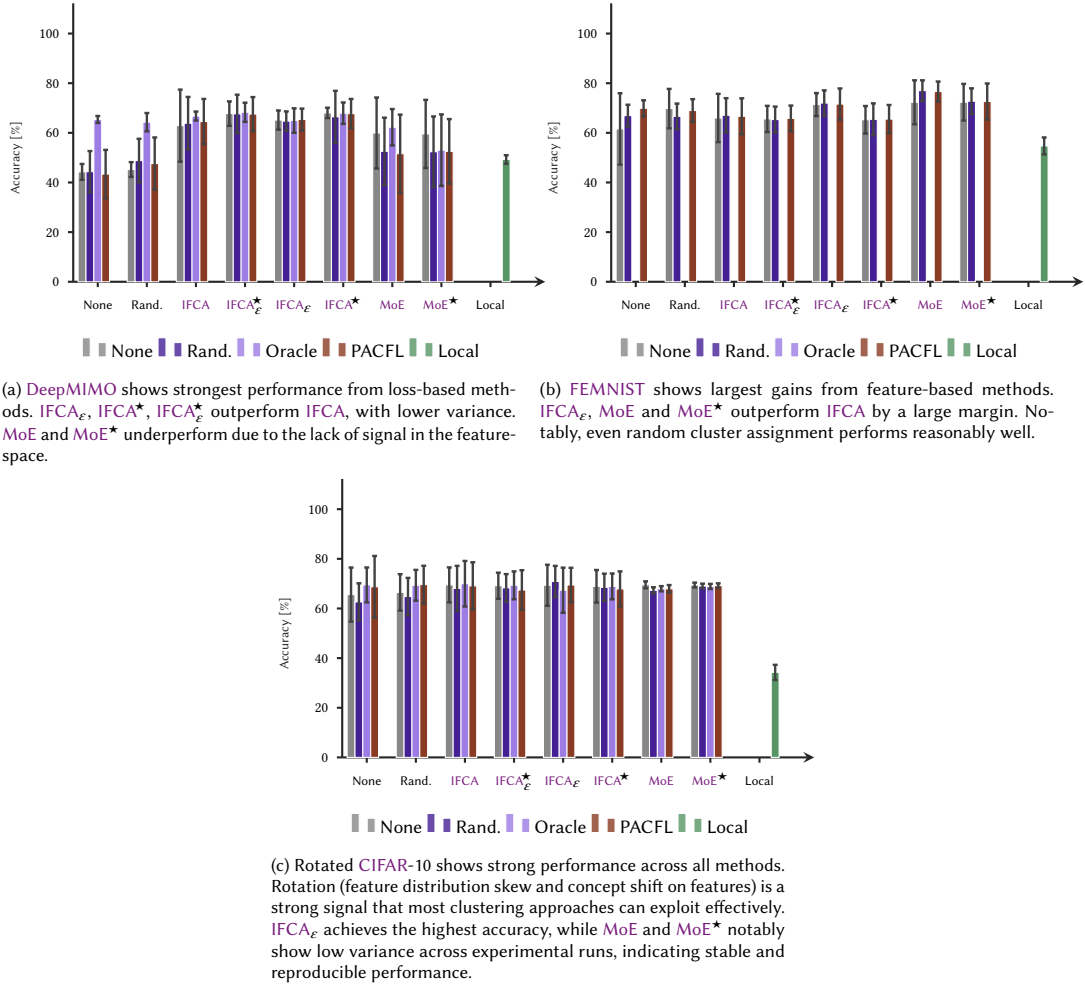

    \centering
    \subfloat[\deepmimo{} shows strongest performance from loss-based methods. \ifcae{}, \ifcap{}, \ifcaep{} outperform \ifca{}, with lower variance. \moe{} and \moes{} underperform due to the lack of signal in the feature-space.]{\includegraphics[width=0.45\linewidth, axisratio=1.78]{data/figures/results_deepmimo_accuracy.tikz}\label{fig:deepmimo_valacc}}%
    \hspace{1em}%
    \subfloat[\femnist{} shows largest gains from feature-based methods. \ifcae{}, \moe{} and \moes{} outperform \ifca{} by a large margin. Notably, even random cluster assignment performs reasonably well.]{\includegraphics[width=0.45\linewidth, axisratio=1.78]{data/figures/results_femnist_accuracy.tikz}\label{fig:femnist_acc}}%
    \hspace{1em}%
    \subfloat[\rotatedcifar{} shows strong performance across all methods. Rotation (feature distribution skew and concept shift on features) is a strong signal that most clustering approaches can exploit effectively. \ifcae{} achieves the highest accuracy, while \moe{} and \moes{} notably show low variance across experimental runs, indicating stable and reproducible performance.]{\includegraphics[width=0.45\linewidth, axisratio=1.78]{data/figures/results_cifar10rot_accuracy.tikz}\label{fig:rotcifar_acc}}%
    \caption{\textbf{Average client accuracy} for different data and model
clustering methods, compared to local training. The x-axis shows the model
clustering method and the bar colour the data-domain cluster initialization.
\emph{None} model clustering method trains a single model per initialized
cluster (\gls{FedAvg} when data-domain clustering is \emph{None}.)
}
    \label{fig:accuracy}
\end{figure*}

\begin{figure*}
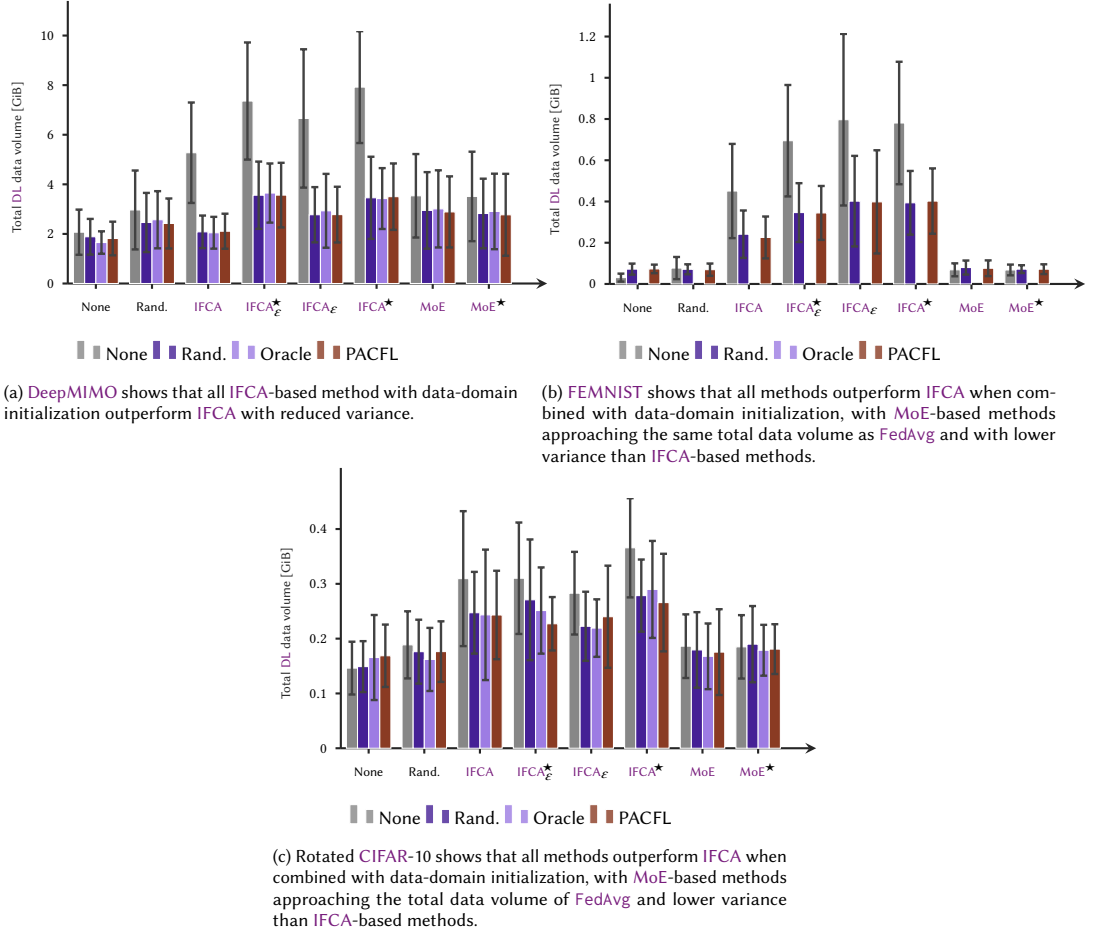

    \centering
    \subfloat[\deepmimo{} shows that all \ifca{}-based method with data-domain initialization outperform \ifca{} with reduced variance.]{\includegraphics[width=0.45\linewidth, axisratio=1.78]{data/figures/results_deepmimo_totaldatadl.tikz}\label{fig:deepmimo_totaldatadl}}%
    \hspace{1em}%
    \subfloat[\femnist{} shows that all methods outperform \ifca{} when combined with data-domain initialization, with \moe{}-based methods approaching the same total data volume as \gls{FedAvg} and with lower variance than \ifca{}-based methods.]{\includegraphics[width=0.45\linewidth,axisratio=1.78]{data/figures/results_femnist_totaldatadl.tikz}\label{fig:femnist_totaldatadl}}%
    \hspace{1em}%
    \subfloat[\rotatedcifar{} shows that all methods outperform \ifca{} when combined with data-domain initialization, with \moe{}-based methods approaching the total data volume of \gls{FedAvg} and lower variance than \ifca{}-based methods.]{\includegraphics[width=0.45\linewidth, axisratio=1.78]{data/figures/results_cifar10rot_totaldatadl.tikz}\label{fig:rotcifar_totaldatadl}}%
    \caption{\textbf{Total \gls{DL} data volume} for different data and model clustering methods. Encoding as in~\cref{fig:accuracy}.}
    \label{fig:totaldatadl}
\end{figure*}

\begin{figure*}
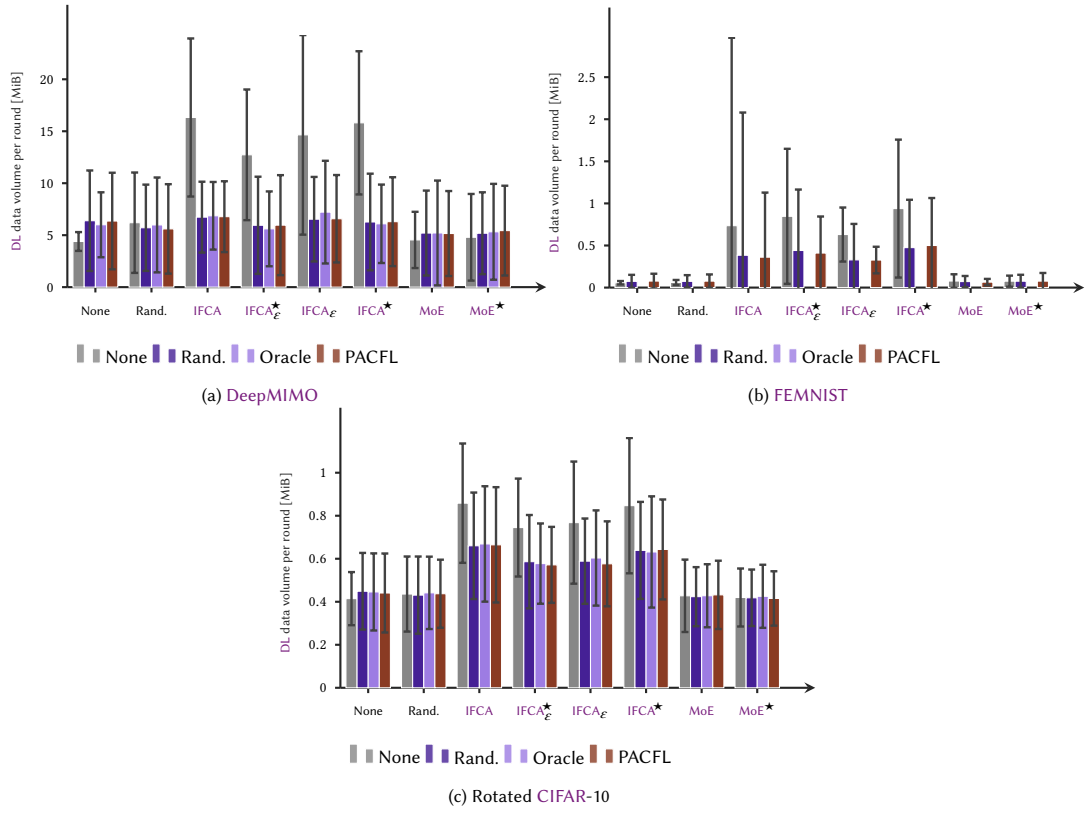

    \centering
    \subfloat[\deepmimo{}]{\includegraphics[width=0.45\linewidth, axisratio=1.78]{data/figures/results_deepmimo_dataperrounddl.tikz}\label{fig:deepmimo_dataperrounddl}}%
    \hspace{1em}%
    \subfloat[\femnist{}]{\includegraphics[width=0.45\linewidth,axisratio=1.78]{data/figures/results_femnist_dataperrounddl.tikz}\label{fig:femnist_dataperrounddl}}%
    \hspace{1em}%
    \subfloat[\rotatedcifar{}]{\includegraphics[width=0.45\linewidth, axisratio=1.78]{data/figures/results_cifar10rot_dataperrounddl.tikz}\label{fig:rotcifar_dataperrounddl}}%
    \caption{\textbf{\gls{DL} data volume per communication round shows consistent patterns.} \ifcae{} and \ifcaep{} reduce per-round communication compared to \ifca{} through \gls{epsgreedy} cluster assignment. \moe{} and \moes{} match \gls{FedAvg} by downloading only a single global cluster model per round, enabling clustering benefits without the communication overhead. Encoding as in~\cref{fig:accuracy}.}
    \label{fig:dataperrounddl}
\end{figure*}

\begin{figure*}
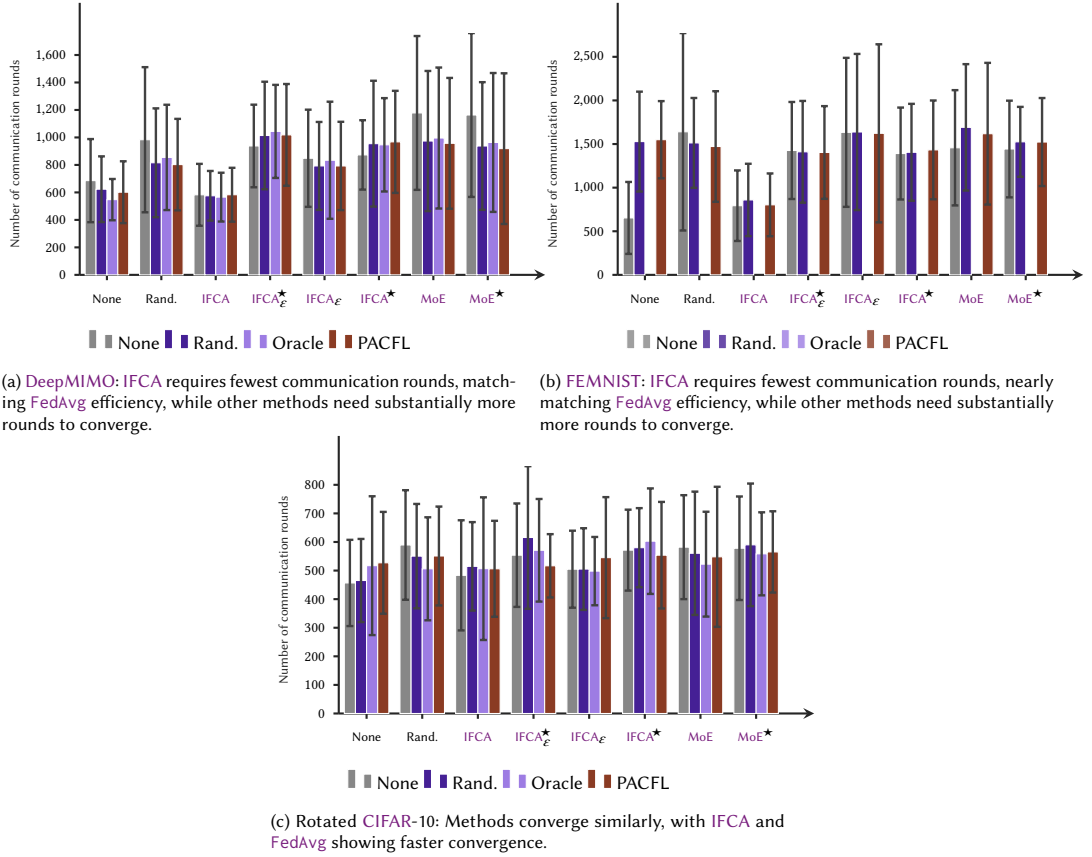

    \centering
    \subfloat[\deepmimo{}: \ifca{} requires fewest communication rounds, matching \gls{FedAvg} efficiency,
while other methods need substantially more rounds to converge.]{\includegraphics[width=0.45\linewidth, axisratio=1.8]{data/figures/results_deepmimo_rounds.tikz}\label{fig:deepmimo_rounds}}%
    \hspace{1em}%
    \subfloat[\femnist{}: \ifca{} requires fewest communication rounds, nearly matching \gls{FedAvg} efficiency, while other methods need substantially more rounds to converge.]{\includegraphics[width=0.45\linewidth,axisratio=1.8]{data/figures/results_femnist_rounds.tikz}\label{fig:femnist_rounds}}%
    \hspace{1em}%
    \subfloat[\rotatedcifar{}: Methods converge similarly, with \ifca{} and \gls{FedAvg} showing faster convergence.]{\includegraphics[width=0.45\linewidth, axisratio=1.8]{data/figures/results_cifar10rot_rounds.tikz}\label{fig:rotcifar_rounds}}%
    \caption{\textbf{Average number of communication rounds} for different data and model clustering methods. Encoding as in~\cref{fig:accuracy}.}
    \label{fig:rounds}
\end{figure*}

\begin{figure*}
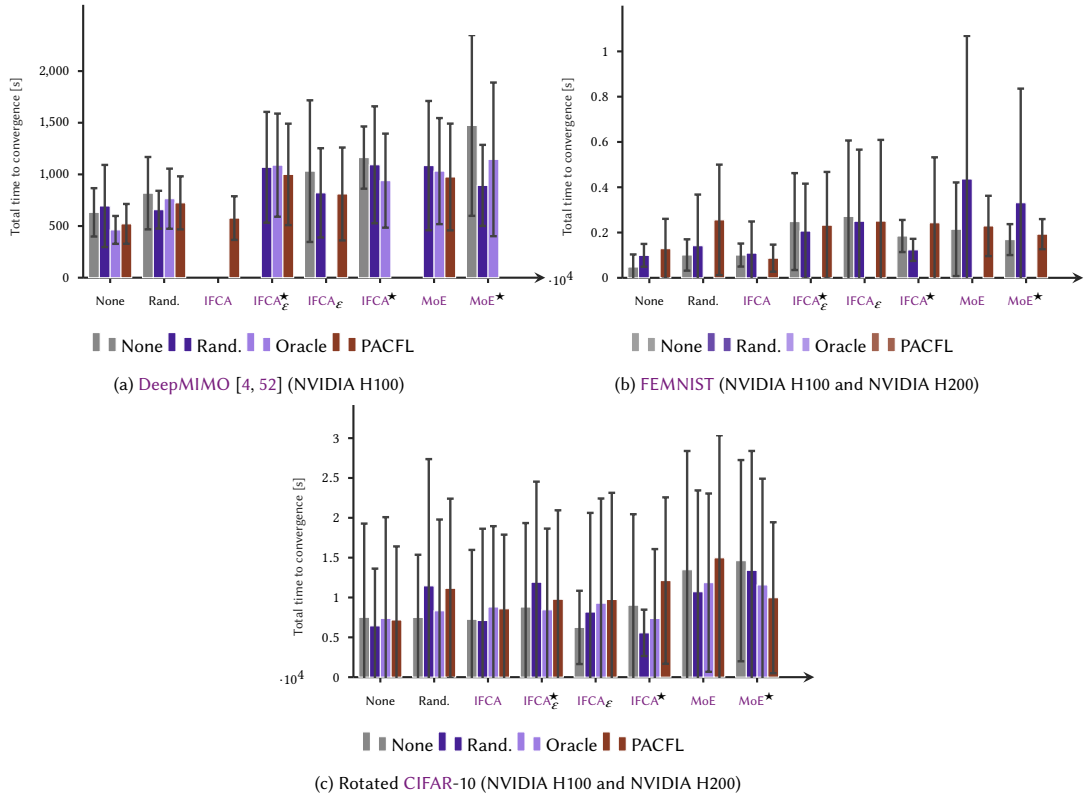

    \centering
    \subfloat[\deepmimo{}~\cite{Remcom,alrabeiah2019deep} (NVIDIA H100)]{\includegraphics[width=0.45\linewidth, axisratio=1.78]{data/figures/results_deepmimo_convergence.tikz}\label{fig:deepmimo_convergence}}%
    \hspace{1em}%
    \subfloat[\femnist{} (NVIDIA H100 and NVIDIA H200)]{\includegraphics[width=0.45\linewidth,axisratio=1.78]{data/figures/results_femnist_convergence.tikz}\label{fig:femnist_convergence}}%
    \hspace{1em}%
    \subfloat[\rotatedcifar{} (NVIDIA H100 and NVIDIA H200)]{\includegraphics[width=0.45\linewidth, axisratio=1.78]{data/figures/results_cifar10rot_convergence.tikz}\label{fig:rotcifar_convergence}}%
    \caption{\textbf{Total time to convergence} for different data and model clustering methods. Note here that we present only data for the faster \glspl{GPU} that we have available. Encoding as in~\cref{fig:accuracy}.}
    \label{fig:conv}
\end{figure*}

\begin{figure*}
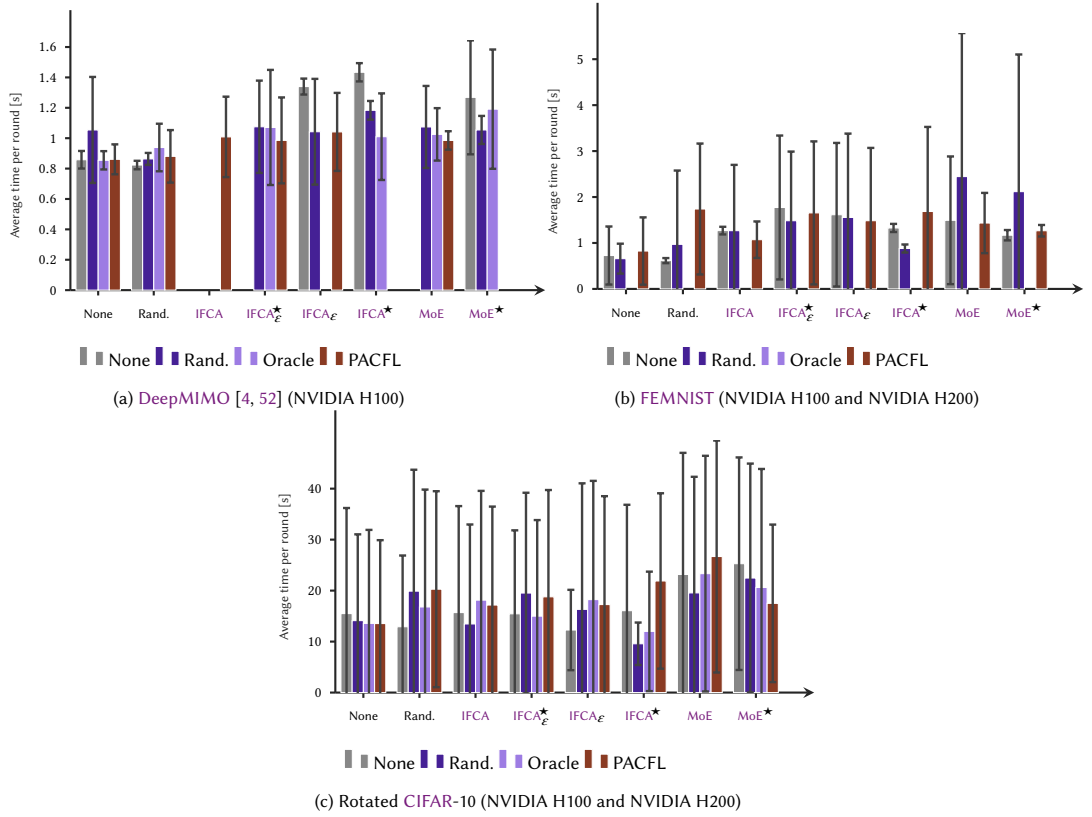

    \centering
    \subfloat[\deepmimo{}~\cite{Remcom,alrabeiah2019deep} (NVIDIA H100)]{\includegraphics[width=0.45\linewidth, axisratio=1.78]{data/figures/results_deepmimo_timeperround.tikz}\label{fig:deepmimo_timeperround}}%
    \hspace{1em}%
    \subfloat[\femnist{} (NVIDIA H100 and NVIDIA H200)]{\includegraphics[width=0.45\linewidth,axisratio=1.78]{data/figures/results_femnist_timeperround.tikz}\label{fig:femnist_timeperround}}%
    \hspace{1em}%
    \subfloat[\rotatedcifar{} (NVIDIA H100 and NVIDIA H200)]{\includegraphics[width=0.45\linewidth, axisratio=1.78]{data/figures/results_cifar10rot_timeperround.tikz}\label{fig:rotcifar_timeperround}}%
    \caption{\textbf{Average time per communication round} for different data and model clustering methods. Note here that we present only data for the faster \glspl{GPU} that we have available (NVIDIA H100 and NVIDIA H200). Encoding as in~\cref{fig:accuracy}.}
    \label{fig:timeperround}
\end{figure*}

%% file: chapters/relatedwork.tex
Non-\glsxtrshort{IID} mitigations come in many forms, such as data sharing
across clients~\cite{zhao2022federated} which reduces distribution divergence
but discloses raw data. Another form of non-\glsxtrshort{IID} mitigation
is~\gls{CFL} which divides clients into clusters where the data distributions
are approximately \glsxtrshort{IID} internally. In general, \gls{CFL} can be
done~\cite{elrifai2025survey} in
\begin{enumerate*}[
  label=(\textit{\roman*}),
  itemjoin={{; }},
  itemjoin*={{; and }},
  after={{.}\ }
]
\item the latent space using model weights or gradients (server-side)
\item using training loss (client-side)
\item using available data or metadata (such as local
data statistics, geographical coordinates, or categorical variables such as
software versions)\end{enumerate*}

\subsection{Models and gradient-based (server-side) clustering}

Server-side~\gls{CFL} methods use the similarity of models or gradients to
cluster clients~\cite{sattler2021clustered,briggs2020federated,long2023multicenter,wang2025wang,
kim2024clustered}. This requires the parameter server to inspect individual client updates, which conflicts with \gls{SA} methods such
as~\cite{bonawitz2017practicalsecure,isaksson2020secure} and complicates \gls{DP}~\cite{abadi2016deep}.

\subsection{Training loss-based (client-side) clustering}

In some studies, clients are presented with a set of possible models and are
tasked with evaluating their own cluster assignment. Examples include
\gls{hypcluster}~\cite{mansour2020three} and \ifca~\cite{ghosh2020ifca},
\ifcae{}~\cite{isaksson2023adaptive} and our method \ifcaep{}. Clients however
need to download and evaluate multiple models, increasing communication and
client complexity~\cite{ghosh2020ifca}. Furthermore, the random
initialization may cause mode collapse~\cite{wu2022motley}. The authors
of~\cite{wu2022motley} note that warm-starting \ifca{} with a few rounds of
\gls{FedAvg} is a partial mitigation to the mode collapse, but that this remains
an open problem. Another possible mitigation to the mode collapse and to the
communication overhead is to initialize cluster assignments. We address these
issues by initializing cluster assignments using data-domain information,
aiming to reduce \gls{DL} data volume and mitigating mode
collapse.

\subsection{Data- and metadata-based clustering}

Clustering can also be done using local data, or available metadata such as
geographical distance between clients (base stations)~\cite{iyer2018mitigating}, or categorical variables such
as software versions or device identifiers~\cite{reisser2021moe}. The telecom
domain is rich in such side-information that is not available in
the open datasets we use here.

As an example of clustering in the data-domain, in
\gls{PACFL}~\cite{vahidian2023efficient} the authors perform clustering in the
data-domain using a low-rank approximation of client data via truncated
\gls{SVD}, see~\cref{sec:data_domain_similarity}.

\gls{FLIS}~\cite{morafah2023flis} uses a similarity measurement calculated from client model predictions on a shared reference dataset to construct a similarity matrix which is used for clustering in a similar way to \gls{PACFL}.

Stenhammar~\etal.~\cite{stenhammar2025clustering} use geographical information
to segment a road network into regions, and jointly optimize cluster assignments
and training to forecast \gls{QoS} for vehicles. This could be considered a hybrid of metadata-based and loss-based approaches.

The authors of \gls{ProCFL}~\cite{xu2025towards} propose a gradient-free
asymmetric similarity metric \gls{ISM} based on label counts per client,
which is used to construct a similarity matrix. This similarity matrix is
clustered using weighted set covering. We expect real-world deployments to
exhibit significant heterogeneity in the feature-space (see~\cref{sec:discussion}). However, combining
feature-based (\gls{PACFL}) and label-based (\gls{ProCFL}) similarity metrics in
order to handle unknown data heterogeneity is an interesting direction for
future work.

\subsection{\glsfmtfull{MoE}}

\citet{zec2020federated} use a \gls{MoE} with a local model and a
global model as experts. \citet{isaksson2023adaptive} improve
on~\cite{zec2020federated} and use a set of trained global cluster models as
input. In this work, we use the set of untrained global cluster models as
experts, and use the gating model to select only a single expert. The gating
model therefore replaces the evaluate step of loss-based methods such as
\ifca{}.

In \gls{fedmix}~\cite{reisser2021moe}, training is conditioned on available
side-information. During training, all experts are initially updated until
specialization is achieved and the speed of this specialization is controlled
via a dampening parameter. They note that dampening has the effect that experts
will learn outside their ``expertise''. Our \moe{} and \moes{} variants share
the idea of gradual specialization, but differ in details. Specifically, in
\moe{} and \moes{}, clients download a single expert model, reducing \gls{DL}
volume compared to \gls{J} models for \gls{fedmix}.

\gls{MoE} has also been used in the vertical \gls{FL} setting, for
example~\cite{folino2024efficientlyapproachingvertical} where a server-side
sparse gating model was used to select one or more experts. This differs from
our setting, where the gating model is local to each client and therefore does
not need any input data to be sent to the server. Another difference is that we
select a single expert to reduce communication. The trade-off between
communication cost savings and local training is worth investigating.

\subsection{Hierarchical \gls{FL}}

\gls{HFL}~\cite{liu2020clientedgecloudhierarchicalfederateda} introduces
multi-tier aggregation to reduce communication latency. In \gls{HFL}, an edge
server aggregates client model updates, and the parameter server in turn
aggregates edge server model updates. This makes \gls{HFL} and \gls{CFL}
composable. Since \gls{HFL} can exploit geographical similarity, combining it with \gls{CFL} is a promising direction for future work..

%% file: chapters/discussion.tex
We find that the best of our methods on each dataset reduces \gls{DL} data
volume while maintaining or improving accuracy over \ifca{} across the evaluated
non-\glsxtrshort{IID} settings. The \moe{} method achieves \qty{10.6}{\pp}
higher accuracy on \femnist{} compared to \ifca{} with \qty{83.1}{\percent}
lower \gls{DL} data volume. Data-domain cluster initialization, \gls{PACFL} or
random, substantially reduces total \gls{DL} data volume, mainly through lower
per-round volume by re-evaluating all cluster models less often, with mixed
effects on communication rounds.

Furthermore, we find that methods that rely on feature-space similarity, such as
\gls{PACFL}, are working well in specific non-\glsxtrshort{IID} scenarios, but
under-performing for \deepmimo{} where the non-\glsxtrshort{IID} characteristics
are most prevalent in the label space. This is expected. In other words, the \gls{sub6} input features are more or less \glsxtrshort{IID} due to the
simulation environment and the similarities between the base stations---with a
notable exception to the third base station, see~\cref{fig:map}. We find that
\ifca{}-based methods perform better here due to the loss-based clustering
capturing this non-\glsxtrshort{IID} characteristic better.

\paragraph{Model capacity} In~\cite{reisser2021moe}, the authors note that
compared to \gls{FedAvg}, the capacity of the expert models can be reduced. We
have not investigated this effect. If the global cluster models can be made
smaller, our \gls{MoE}-based methods can potentially reduce the \gls{DL} data
volume to lower than that of \gls{FedAvg}.

\paragraph{Number of cluster models}

The number of cluster models is a hyperparameter.
\cite{isaksson2023adaptive,isaksson2023mmwave}~showed that \ifca{}-based methods
are robust to setting the number of cluster models too high, with the drawback
of increased communication. Our methods decrease the per communication round
\gls{DL} data volume, with \moe{}-based methods reducing per-round \gls{DL} data
volume to that of the baseline \gls{FedAvg}.

\paragraph{Public datasets}

There are few open public datasets in the telecom
domain which can be directly used for evaluating distributed
algorithms. The realistic datasets needed to properly
evaluate large scale distributed algorithms and frameworks such as those
proposed here are generally not available for researchers. For example, we have
not made use of side-information that can be used for similarity estimation in
the telecom domain, such as geographical locations, configuration parameters or
performance indicators. This remains unexplored.

\paragraph{Generalization}

While priming our methods with an initial estimate of cluster membership does not
significantly change accuracy, it does have an effect on reducing \gls{DL} data
volume, driven by per-round volume. When deployed in a real scenario, the system
will have to continually adapt to new and changing non-\glsxtrshort{IID}
distributions and generalize to new unseen clients~\cite{elrifai2025survey}.
These aspects are underexplored in the telecom domain.

We assume that clients are stationary. \gls{PACFL}~\cite{vahidian2023efficient}
includes an algorithm to determine cluster membership of the new clients
entering the system without changing the cluster membership of existing clients.

\paragraph{Radio environment}

\deepmimo{} is a simplified simulation environment that may not capture
real-world channel complexity. For example, in this work we have assumed a
horizontal \gls{ULA}, while most actual products use rectangular \gls{URA} in
order to fit sufficient number of antennas in constrained spaces. Practical
deployments favor vertical configurations.

Our beams are defined based on a beam angle, while most practical deployments
use \gls{FFT} beams. We argue that the difference for the purpose of this paper
is small.

The non-\glsxtrshort{IID} characteristics seen in the~\deepmimo{} dataset stem from three causes.

\begin{enumerate*}[
  label=(\textit{\roman*}),
  itemjoin={{; }},
  itemjoin*={{; and }},
  after={{.}\ }
]
    \item The environment in \deepmimo{} Outdoor~\num{1}
Blockage scenario consists of a road (which is where \glspl{UE} are located), buildings of different height~\cite{Remcom,deepmimov23, DBLP:journals/corr/abs-1902-06435} which reflect the beams and give rise to \gls{UE} spatial distributions
    \item Blockage and reflectors in front of a single antenna cause a non-uniform distribution of \glspl{UE}
    \item The mirror property of the \gls{ULA} antenna (\cref{fig:ula}) is used to construct two clusters of clients, as seen in~\cref{fig:map}
\end{enumerate*}

The non-\glsxtrshort{IID} characteristics stem more from \gls{UE} distributions
than differences in the channel characteristics between base stations, and our
hypothesis is that data distributions between sites in real life will come from
differences in channel characteristics to a higher degree. Furthermore, the
weakest \glspl{UE} often need beamforming the most, something not considered in
the clustering nor evaluated in this paper.

We use accuracy as a performance metric for all of our datasets, including
\deepmimo{}. This may not correspond to actual performance in real networks as
users can be served from sidelobes of other beams with good performance. This is
further examined in~\cite{isaksson2023mmwave,cheng2022}.

We therefore argue that a realistic dataset would contain non-\glsxtrshort{IID}
characteristics from a wider set and that our combined datasets show that our
methods perform well in different scenarios. We include \femnist{} and
\rotatedcifar{} to demonstrate the effectiveness of our methods across different
non-\glsxtrshort{IID} types, but these do not validate wireless-specific claims.
It is expected that \gls{MoE}-based methods excel in real-world deployments with
stronger feature-space heterogeneity between base stations. Real-world
validation is essential.

\paragraph{Number of clients}

Scalability to thousands and tens of thousands of clients remains to be studied.
New and larger, telecom specific benchmarks are important to reach this goal.

\subsection{Privacy}

\gls{CFL} can reveal similarities between clients through cluster
assignments~\cite{kim2023exploring,xu2025towards}.
\gls{PACFL}~\cite{vahidian2023efficient} uses truncated \gls{SVD} which is
known to reveal the dominant directions of local data distributions. Mitigations
including \gls{SA}~\cite{isaksson2020secure} and
\gls{DP}~\cite{kim2023exploring,xu2025towards} are important directions of
future work.

\subsection{Personalized \gls{FL}}

A common method for addressing non-\glsxtrshort{IID}-ness in \gls{FL} is using
methods to limit the local drift from a global model---such as
\gls{fedprox}~\cite{li2020federated} which adds a proximal term to penalize
deviation from the global model and
\gls{scaffold}~\cite{karimireddy2020scaffold} which estimates and controls the
drift between local and global gradients. Such methods can be applied directly
in our framework and studying their interactions is another direction of future
work.

\subsection{Hyperparameters}

Hyperparameters were tuned individually per dataset
using~\cite{liaw2018tune}, see~\cref{sec:training,sec:hyperparams}.
However, there are many hyperparameters interacting with each other in
interesting ways, and their effects should be studied more closely.

%% file: chapters/conclusion.tex
We proposed a modular \gls{CFL} framework designed
to operate efficiently under non-\glsxtrshort{IID} data distributions typical of heterogeneous wireless environments. Our approach combines three complementary components:\begin{enumerate*}[
  label=(\textit{\roman*}),
  itemjoin={{; }},
  itemjoin*={{; and }},
  after={{.}\ }
]
    \item data-domain cluster initialization using principal-angle similarity
    \item model-based clustering during training with loss-proportional cluster selection
    \item a communication-efficient \gls{MoE} formulation with exploration mechanisms
\end{enumerate*}

The proposed framework addresses two central challenges in \gls{FL} for \gls{6G}
systems: performance degradation under non-\glsxtrshort{IID} data and excessive
communication overhead in \gls{CFL} training. Specifically, integrating
\gls{CFL} and \gls{MoE} models can improve performance under
non-\glsxtrshort{IID} settings for \gls{mmWave} beam
selection~\cite{isaksson2023mmwave,cheng2022}. However, these methods suffer
from high communication overhead and computational complexity, limiting their
scalability in large-scale \gls{6G} systems.

By combining data-domain cluster initialization with adaptive model-based
clustering, we improve personalization without sharing raw data, and local
gating models remain entirely on-device. Across three datasets exhibiting
diverse non-\glsxtrshort{IID} characteristics---\deepmimo{} (concept shift on
labels and quantity skew), \femnist{} (feature distribution skew, label
distribution skew, and quantity skew), and \rotatedcifar{} (feature distribution
skew and concept shift on features)---the best method for each dataset matches
or improves performance compared to standard \ifca{} and \gls{FedAvg} baselines.
No single method is superior on all datasets. In particular, \moe{} with
\gls{PACFL} improves accuracy by \qty{10.6}{\pp} while reducing total \gls{DL}
data volume by \qty{83.1}{\percent} relative to \ifca{} on \femnist{}. \ifcap{}
with \gls{PACFL} improves accuracy by \qty{4.8}{\pp} on \deepmimo{}.

A key insight from our study is that different non-\glsxtrshort{IID}
characteristics favor different clustering strategies. In our experiments,
feature-space similarity is particularly effective under covariate shift
(\femnist), whereas model-based clustering provides greater benefits under
concept shift on labels (\deepmimo{}). Under feature distribution skew combined with concept shift on features (\rotatedcifar{}), no method significantly improves accuracy over \ifca{}, but
\moe{} with \gls{PACFL} still reduces total \gls{DL} data volume by
\qty{43.3}{\percent}. The proposed framework allows these mechanisms to be
combined in a modular manner, enabling adaptation to diverse deployment
scenarios.

While our evaluation is conducted on a simulation dataset and
established benchmarks, several directions remain for future work. These include
establishing formal convergence guarantees for the \gls{MoE}-based clustering
mechanism, extending the framework to larger, highly dynamic client populations,
extending to a continual learning scenario, and evaluating performance under
real-world telecom deployment data. Further investigation into adaptive
selection of the number of clusters and expert capacity reduction may yield
additional communication gains. Overall, this work shows that adaptive and
communication-aware clustering is a promising direction for scalable,
resource-efficient federated learning in future \gls{6G} networks.

%% file: includes/acknowledgment.tex
This work was partially supported by the \gls{WASP}
funded by the Knut and Alice Wallenberg Foundation.%

During the preparation of this work, the authors used \gls{AI} coding assistants
(Amazon~Q Developer with Claude Sonnet~4.5; Anthropic Claude Code with Claude
Opus~4.7--5) for implementing statistical tests, code refactoring,
editorial review, and consistency checks. The authors reviewed and edited the
content as needed and take full responsibility for the content.

%% file: chapters/appendix.tex
\section{Methods}\label{sec:appmethods}
\subsection{\glsfmtfull{IFCA}}\label{sec:appendix_ifca}

This section details \ifca{} including our additions. See~\cref{alg:fl.server,alg:fl.preclustering,alg:fl.truncsvd,alg:fl.client,alg:fl.cluster.ifca,alg:fl.cluster.ifcap,alg:fl.update}.

Client-level accuracy distributions are shown in~\cref{fig:accuracycdf}.

\begin{algorithm*}[tbh]
      \begin{algorithmic}[1]
      \Procedure{server}{$\gls{C},\gls{K}$}
          \State \hlf{$\gls{J} \gets \Call{cluster}{\gls{principalcomponents}, \gls{S}}$}\hfill\Comment{Cluster clients according to similarity matrix.}
          \State$\text{initialize}\,\gls{Jremset} \gets \gls{Jset}, \left\{\gls{wgjt0} \given \gls{j} \in \gls{Jremset}\right\}$
          \State$\gls{Ks} \gets \left\lceil \gls{C}\gls{K} \right\rceil$\hfill\Comment{\parbox[t]{.45\linewidth}{Number of clients to select per communication round}}
          \For{$\gls{t} \in \{1,2, \ldots \}$}\Comment{Until convergence}
              \State$\gls{St} \subseteq \gls{S}, \vert\gls{St}\vert = \gls{Ks}$\hfill\Comment{Random sampling of \gls{Ks} clients}
              \ForAll{$\gls{k} \in \gls{St}$}\Comment{For all clients, in parallel}
                   \State$\gls{wkt1}, \gls{nk}, \gls{jhat}^{\gls{k}} \gets \gls{k}.\flconstfnt{client}\left(\left\{\gls{wgj} \given \gls{j} \in \gls{Jremset}\right\}, \gls{t}\right)$\Comment{Local training (\cref{alg:fl.client})}
              \EndFor%
              \ForAll{$\gls{j} \in \gls{Jset}$}\hfill\Comment{For all cluster models}
                  \State$\gls{nj} \gets \sum_{\gls{k} \in \left\{\gls{kp} \in \gls{St} \given \gls{jhat}^{\gls{kp}} = \gls{j}\right\}} \gls{nk}$\hfill\Comment{\parbox[t]{.4\linewidth}{Total number of samples for cluster model \gls{j} from clients where $\gls{j} = \gls{jhat}$}}
                  \State$\gls{wgjt1} \gets \sum_{\gls{k} \in \left\{\gls{kp} \in \gls{St} \given \gls{jhat}^{\gls{kp}} = \gls{j}\right\}} \frac{\gls{nk}}{\gls{nj}}\gls{wkt1}$\hfill\Comment{Update cluster model \gls{j} with clients where $\gls{j} = \gls{jhat}$}
              \EndFor%
          \EndFor%
      \EndProcedure%
      \algstore{bkbreak1}
      \end{algorithmic}
\caption{Data-domain cluster initialization \gls{FL}---server}\label{alg:fl.server}
\end{algorithm*}%
\begin{algorithm}[tbh]
      \begin{algorithmic}[1]
            \algrestore{bkbreak1}
            \Procedure{cluster}{\gls{principalcomponents}, \gls{S}}\Comment{Server data-domain cluster initialization}
            \ForAll{$\gls{k} \in \gls{S}$}\Comment{For all clients, in parallel}
                  \State $\gls{U}_{\gls{k},\gls{principalcomponents}} \gets \Call{truncsvd}{\gls{Mk},\gls{principalcomponents}}$
            \EndFor%
            \ForAll{$\gls{k} \in \gls{S}$}
                  \ForAll{$\gls{kp} \in \gls{S}$}
                        \State $\gls{adjacencymatrix}_{\gls{k}\gls{kp}} \gets \arccos\left(\gls{U}_{\gls{k},\gls{principalcomponents}}^\transpose\gls{U}_{\gls{kp},\gls{principalcomponents}}\right)$\Comment{Compute the angle between two rows of \gls{U} matrices}
                  \EndFor%
            \EndFor%
            \EndProcedure%
            \algstore{bkbreak21}
      \end{algorithmic}
      \caption{data-domain cluster initialization---server~\cite{morafah2022rethinking} (see~\cref{sec:data_domain_similarity})}\label{alg:fl.preclustering}
\end{algorithm}%

\begin{algorithm}[tbh]
      \begin{algorithmic}[1]
            \algrestore{bkbreak21}
            \Procedure{truncsvd}{$\gls{Mk}$, \gls{principalcomponents}}\Comment{Client}\
            \State $\gls{Mk} \approx \gls{U}_{\gls{principalcomponents}}\gls{singularvaluematrix}_{\gls{principalcomponents}}\gls{rightsingularvaluematrix}_{\gls{principalcomponents}}^\transpose$\hfill\Comment{\parbox[t]{.4\linewidth}{Perform Truncated \gls{SVD} of local dataset}}
            \State\textbf{return} $\gls{U}_{\gls{principalcomponents}}$
            \EndProcedure%
            \algstore{bkbreak2}
      \end{algorithmic}
      \caption{Truncated \gls{SVD}---client~\cite{morafah2022rethinking} (see~\cref{sec:data_domain_similarity})}\label{alg:fl.truncsvd}
\end{algorithm}%
\begin{algorithm}[tbh]
      \begin{algorithmic}[1]
      \algrestore{bkbreak2}
      \Procedure{client}{$\left\{\gls{wgj} \given \gls{j} \in \gls{Jremset}\right\}, \gls{t}$}
          \State$\gls{jhat} \gets \flconstfnt{cl.-est.}\left(\gls{eps}, \gls{epssim}, \left\{\gls{wgj} \given \gls{j} \in \gls{Jremset}\right\}\right),\, \gls{nk} \gets \vert\gls{Pk}\vert$\Comment{\parbox[t]{.4\linewidth}{Estimate best cluster (\cref{alg:fl.cluster.ifca} for \ifca{}/\ifcae{}; \cref{alg:fl.cluster.ifcap} for \ifcap{}/\ifcaep{})}}
          \State$\gls{wkt1} \gets \flconstfnt{update}\left(\gls{wgjht}, \gls{nk}\right)$\Comment{\parbox[t]{.4\linewidth}{Perform local training using cluster model~\gls{jhat} (\cref{alg:fl.update})}}
          \State\textbf{return} $(\gls{wkt1}, \gls{nk}, \gls{jhat})$
      \EndProcedure%
      \algstore{bkbreak3}
      \end{algorithmic}
      \caption{\ifca{}-based methods---client~\cite{isaksson2023adaptive,ghosh2020ifca} (see~\cref{sec:ifcap})}\label{alg:fl.client}
\end{algorithm}%
\begin{algorithm}[tbh]
      \begin{algorithmic}[1]
      \algrestore{bkbreak3}
      \Procedure{cl.-est.}{$\gls{eps}, \gls{epssim}, \left\{\gls{wgj} \given \gls{j} \in \gls{Jremset}\right\}$}
          \If{with probability $\gls{eps}$}
              \State \Return $\gls{jhat} \sim \mathcal{U}\left\{1, \gls{J}\right\}$ \Comment{Random exploration}
          \ElsIf{with probability $\gls{epssim}$}
              \State \Return $\gls{jhat} \gets \argmin_{\gls{j} \in \gls{Jremset}} \sum_{i \in \gls{Pk}} \gls{l}\left(\gls{xi}, \gls{yi}, \gls{wgj}\right)$ \Comment{Argmin-loss}
          \Else
              \State \Return previously-assigned $\gls{jhat}$ for this client
          \EndIf
      \EndProcedure%
      \algstore{bkbreak3b}
      \end{algorithmic}
      \caption{Cluster assignment for \ifca{} and \ifcae{}~\cite{ghosh2020ifca,isaksson2023adaptive}}\label{alg:fl.cluster.ifca}
\end{algorithm}%
\begin{algorithm}[tbh]
      \begin{algorithmic}[1]
      \algrestore{bkbreak3b}
      \Procedure{cl.-est.}{$\gls{eps}, \gls{epssim}, \left\{\gls{wgj} \given \gls{j} \in \gls{Jremset}\right\}$}
          \If{with probability $\gls{eps}$}
              \State \Return $\gls{jhat} \sim \mathcal{U}\left\{1, \gls{J}\right\}$ \Comment{Random exploration}
          \ElsIf{with probability $\gls{epssim}$}
              \State $\gls{bl} \gets \left[\left(\sum_{i \in \gls{Pk}} \gls{l}\left(\gls{xi}, \gls{yi}, \gls{wgj}\right)\right)^{-1}\right]_{\gls{j} \in \gls{Jremset}}$
              \State \Return $\gls{jhat}$ sampled from $\gls{Jremset}$ with probability $\propto \gls{bl}$ \Comment{Loss-proportional}
          \Else
              \State \Return previously-assigned $\gls{jhat}$ for this client
          \EndIf
      \EndProcedure%
      \algstore{bkbreak4}
      \end{algorithmic}
      \caption{Cluster assignment for \ifcap{} and \ifcaep{}}\label{alg:fl.cluster.ifcap}
\end{algorithm}%
\begin{algorithm}[tbh]
      \begin{algorithmic}[1]
      \algrestore{bkbreak4}
      \Procedure{update}{$\gls{wkt1}, \gls{nk}$}\Comment{Mini-batch gradient descent}
          \For{$e \in \gls{Eset}$}\Comment{For a few epochs}
              \ForAll{batches of size \gls{B}}\Comment{Batch update}
                  \State\LComment{Local parameter update}
                  \State$\gls{wkt1} \gets \gls{wkt1} - \frac{\gls{eta}}{\gls{B}} \nabla_{\gls{wkt1}} \sum_{i=1}^{\gls{B}} \gls{l}\left(\gls{xi}, \gls{yi}, \gls{wkt1}\right)$w
              \EndFor%
          \EndFor%
          \State\textbf{return} $\gls{wkt1}$
      \EndProcedure%
      \end{algorithmic}
      \caption{\gls{FL}---local update~\cite{mcmahan2017communication}}\label{alg:fl.update}
\end{algorithm}

\subsection{\glsfmtfull{MoE}}\label{sec:appendix_moe}

We add an entropy regularization term weighted by \gls{moealpha} to the
loss function. During training, the selected expert gate logit receives a
primary loss gradient and is sharpened through the entropy regularizer. The outer
mechanism (\gls{epsgreedy} or \ucbone) forces exploration of experts.

\begin{equation}
\gls{loss}_{\text{total}} = \gls{loss}_{\text{primary}} + \gls{moealpha} \cdot \gls{loss}_{\text{entropy}},
\end{equation}
where%
\begin{align}
\gls{gateprob} &= \softmax\left(\gls{gatek}(\gls{x})\right),\\
\gls{jhat} &= \argmax_{\gls{j} \in \gls{Jremset}} \left\{\begin{array}{ll}
    \gls{gateprob}_{\gls{j}},&\text{\moe}\\
    \gls{gateprob}_{\gls{j}} + \sqrt{\frac{\gls{ucbconst} \ln \gls{totalclusterusage}}{\gls{clusterusage}}},&\text{\moes}
\end{array}\right.\\
\gls{loss}_{\text{primary}} &= -\log \gls{P}\left(\gls{jhat}, \gls{y} \given \gls{x}, \gls{wgjht}, \gls{gatek}\right),\\
\gls{loss}_{\text{entropy}} &= -\sum_{\gls{j}=1}^{\gls{J}} \gls{gateprob}_{\gls{j}} \log \gls{gateprob}_{\gls{j}},\\
\log \gls{P}\left(\gls{jhat}, \gls{y} \given \gls{x}, \gls{wgjht}, \gls{gatek}\right) &= \log \gls{gateprob}_{\gls{jhat}} + \log f_{\gls{jhat}}(\gls{x})_{\gls{y}},
\end{align}
where \(\gls{P}\left(\gls{jhat}, \gls{y} \given \cdot\right)\) is the joint
probability output for selected expert $\gls{jhat}$. $\gls{wgjht}$ are the
parameters at time \gls{t} of the selected expert $\gls{jhat}$,
$\gls{gatek}$ is the gating model on client \gls{k} and
$f_{\gls{jhat}}(\gls{x})$ is the output of the selected expert.

\section{Detailed results}\label{sec:appresults}

\begin{figure*}
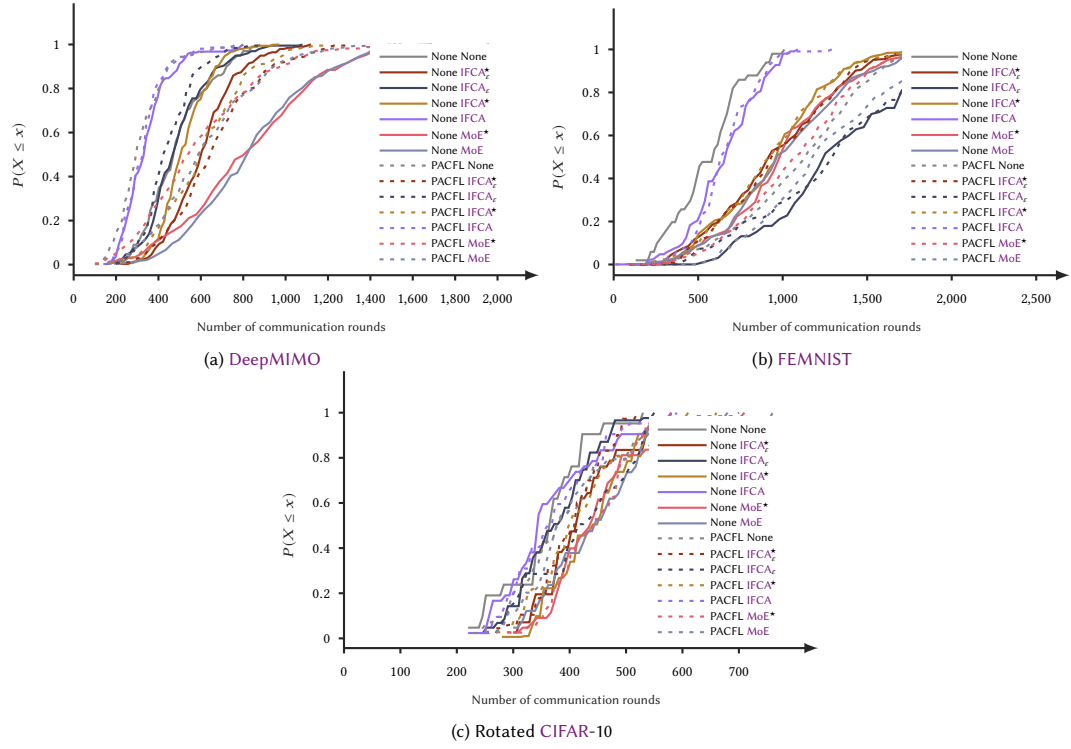

    \centering
    \subfloat[\deepmimo{}]{\includegraphics[width=0.45\linewidth, axisratio=1.8]{data/figures/results_deepmimo_roundscdf.tikz}\label{fig:deepmimo_roundscdf}}%
    \hspace{1em}%
    \subfloat[\femnist{}]{\includegraphics[width=0.45\linewidth,axisratio=1.8]{data/figures/results_femnist_roundscdf.tikz}\label{fig:femnist_roundscdf}}%
    \hspace{1em}%
    \subfloat[\rotatedcifar{}]{\includegraphics[width=0.45\linewidth, axisratio=1.8]{data/figures/results_cifar10rot_roundscdf.tikz}\label{fig:rotcifar_roundscdf}}%
    \caption{\textbf{\gls{CDF} of number of communication rounds} for different data and model clustering methods. Legend entries read \emph{initialization model clustering}, \emph{None None} is \gls{FedAvg}.}
    \label{fig:roundscdf}
\end{figure*}

\begin{figure*}
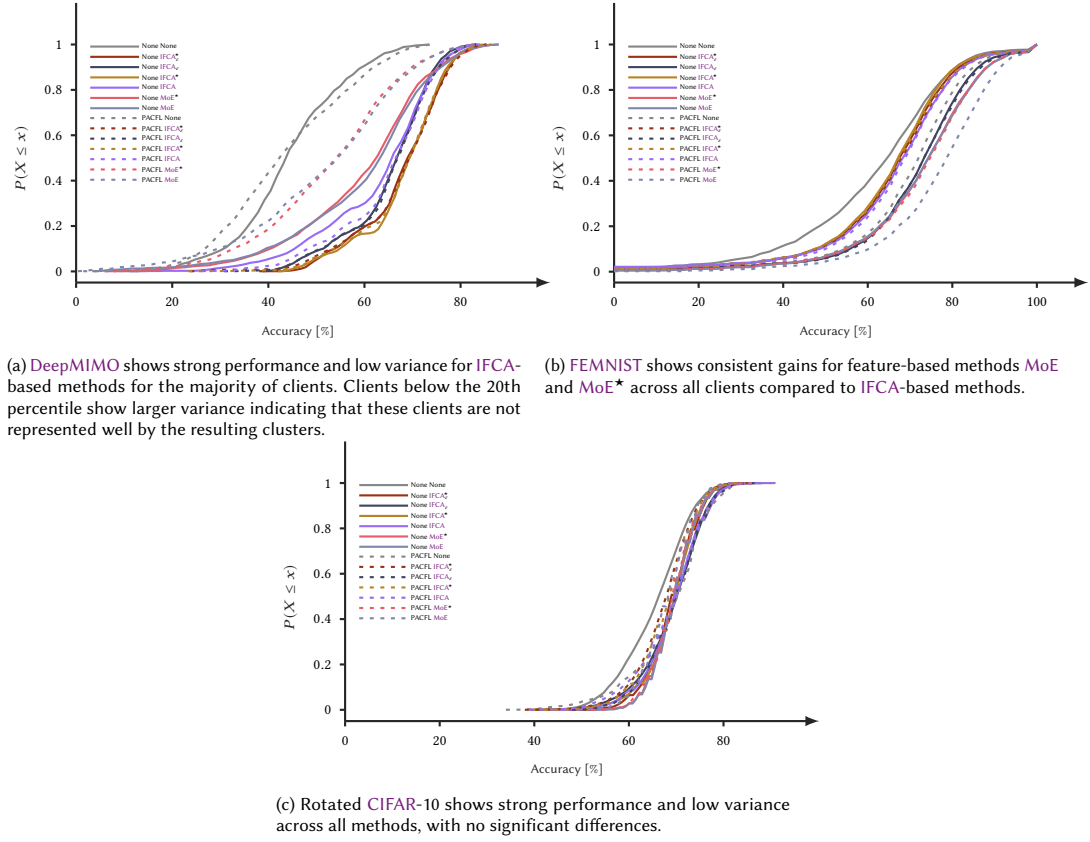

    \centering
    \subfloat[\deepmimo{} shows strong performance and low variance for \ifca{}-based methods for the majority of clients. Clients below the 20th percentile show larger variance indicating that these clients are not represented well by the resulting clusters.]{\includegraphics[width=0.45\linewidth, axisratio=1.78]{data/figures/results_deepmimo_accuracycdf.tikz}\label{fig:deepmimo_valacc_cdf}}%
    \hspace{1em}%
    \subfloat[\femnist{} shows consistent gains for feature-based methods \moe{} and \moes{} across all clients compared to \ifca{}-based methods.]{\includegraphics[width=0.45\linewidth, axisratio=1.78]{data/figures/results_femnist_accuracycdf.tikz}\label{fig:femnist_acc_cdf}}%
    \hspace{1em}%
    \subfloat[\rotatedcifar{} shows strong performance and low variance across all methods, with no significant differences.]{\includegraphics[width=0.45\linewidth, axisratio=1.78]{data/figures/results_cifar10rot_accuracycdf.tikz}\label{fig:rotcifar_acc_cdf}}%
    \caption{\textbf{Client accuracy distribution} for different data and model clustering methods. Legend entries read \emph{initialization model clustering}, \emph{None None} is \gls{FedAvg}.}
    \label{fig:accuracycdf}
\end{figure*}

\begin{figure*}
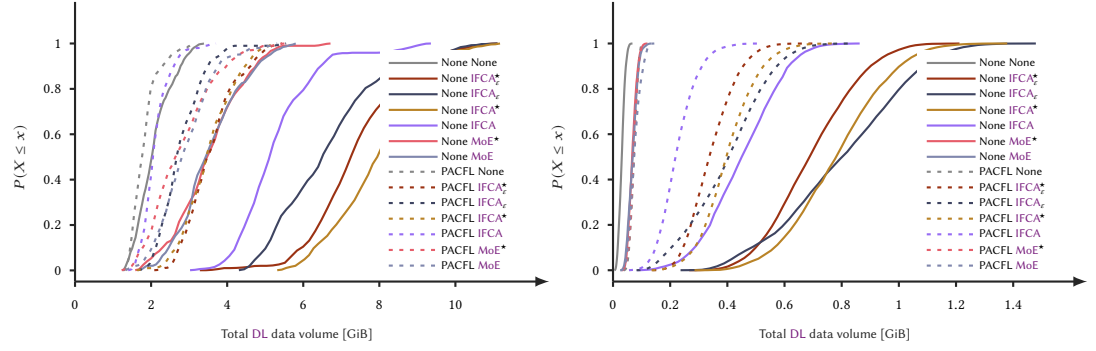
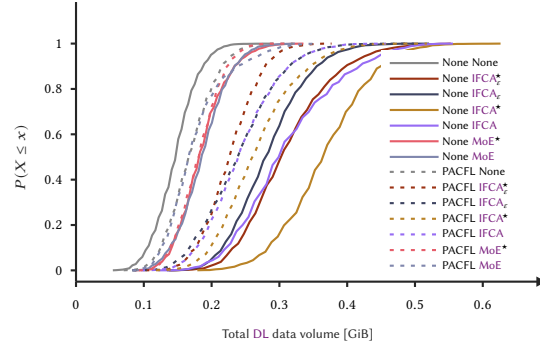

    \centering
    \subfloat[\deepmimo{} shows that all methods with data-domain initialization outperform \ifca{} with reduced variance compared to \ifca{}-based methods. We see the biggest reduction in \gls{DL} data volume for the top percentiles.]{\includegraphics[width=0.45\linewidth, axisratio=1.78]{data/figures/results_deepmimo_totaldatadlcdf.tikz}\label{fig:deepmimo_totaldatadlcdf}}%
    \hspace{1em}%
    \subfloat[\femnist{} shows that all methods with data-domain initialization outperform \ifca{}, with \moe{}-based methods approaching the same total data volume as \gls{FedAvg} and with lower variance than \ifca{}-based methods.]{\includegraphics[width=0.45\linewidth,axisratio=1.78]{data/figures/results_femnist_totaldatadlcdf.tikz}\label{fig:femnist_totaldatadlcdf}}%
    \hspace{1em}%
    \subfloat[\rotatedcifar{} shows that all methods with data-domain initialization outperform \ifca{}, with \moe{}-based methods achieving nearly the same total data volume as \gls{FedAvg} and lower variance than \ifca{}-based methods.]{\includegraphics[width=0.45\linewidth, axisratio=1.78]{data/figures/results_cifar10rot_totaldatadlcdf.tikz}\label{fig:rotcifar_totaldatadlcdf}}%
    \caption{\textbf{\glsxtrshort{CDF} of the total \gls{DL} data volume} for different data and model clustering methods. Legend entries read \emph{initialization model clustering}, \emph{None None} is \gls{FedAvg}.}
    \label{fig:totaldatadlcdf}
\end{figure*}

\input{chapters/appendix_deepmimo.tex}
\input{chapters/appendix_cifar10rot.tex}
\input{chapters/appendix_femnist.tex}

\input{chapters/hyperparams.tex}

%% file: chapters/appendix_deepmimo.tex
\subsection{Detailed results for \deepmimo}
Here follows detailed results for the \deepmimo{} dataset, see~\cref{sec:datasets}.

\begin{figure*}
    \centering
    \subfloat[Base station 12]{\includegraphics[width=0.15\linewidth]{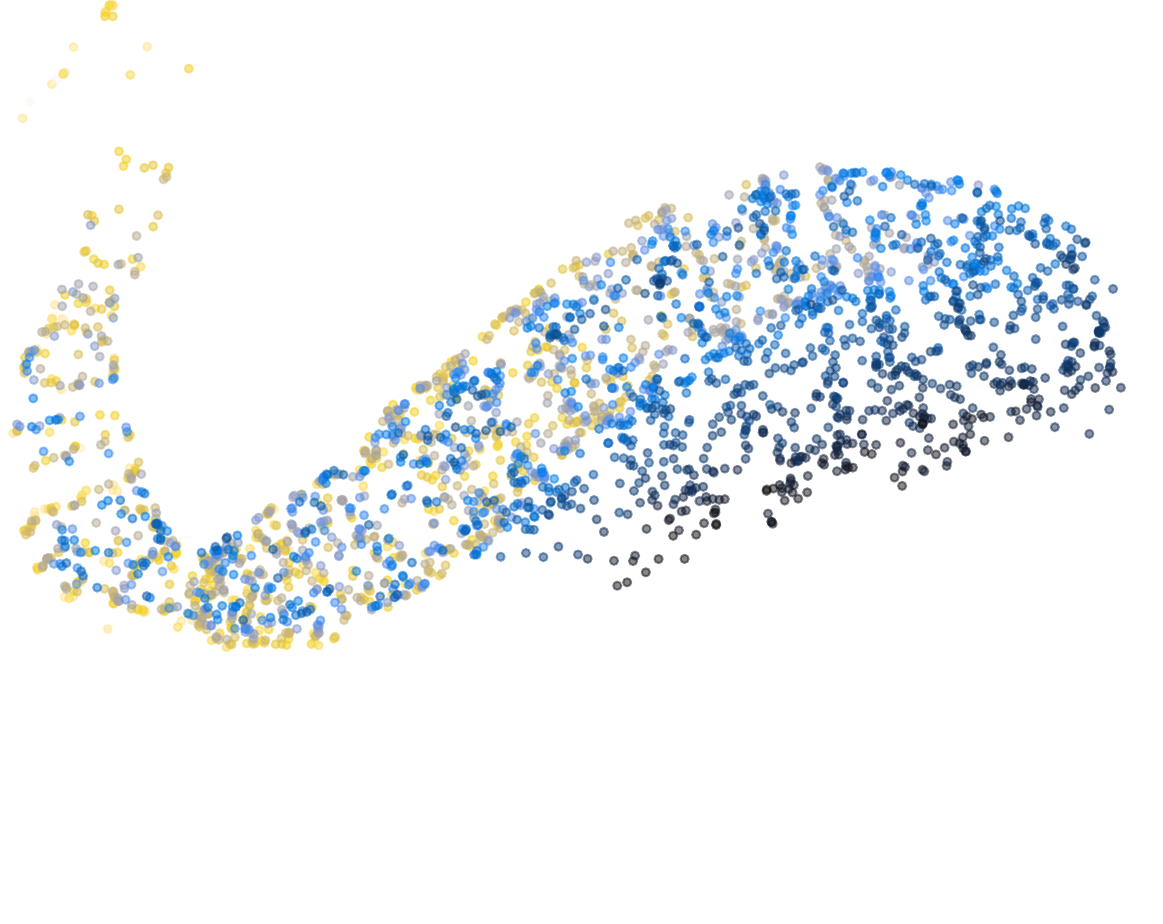}}%
    \hspace{.2cm}%
    \subfloat[Base station 10]{\includegraphics[width=0.15\linewidth]{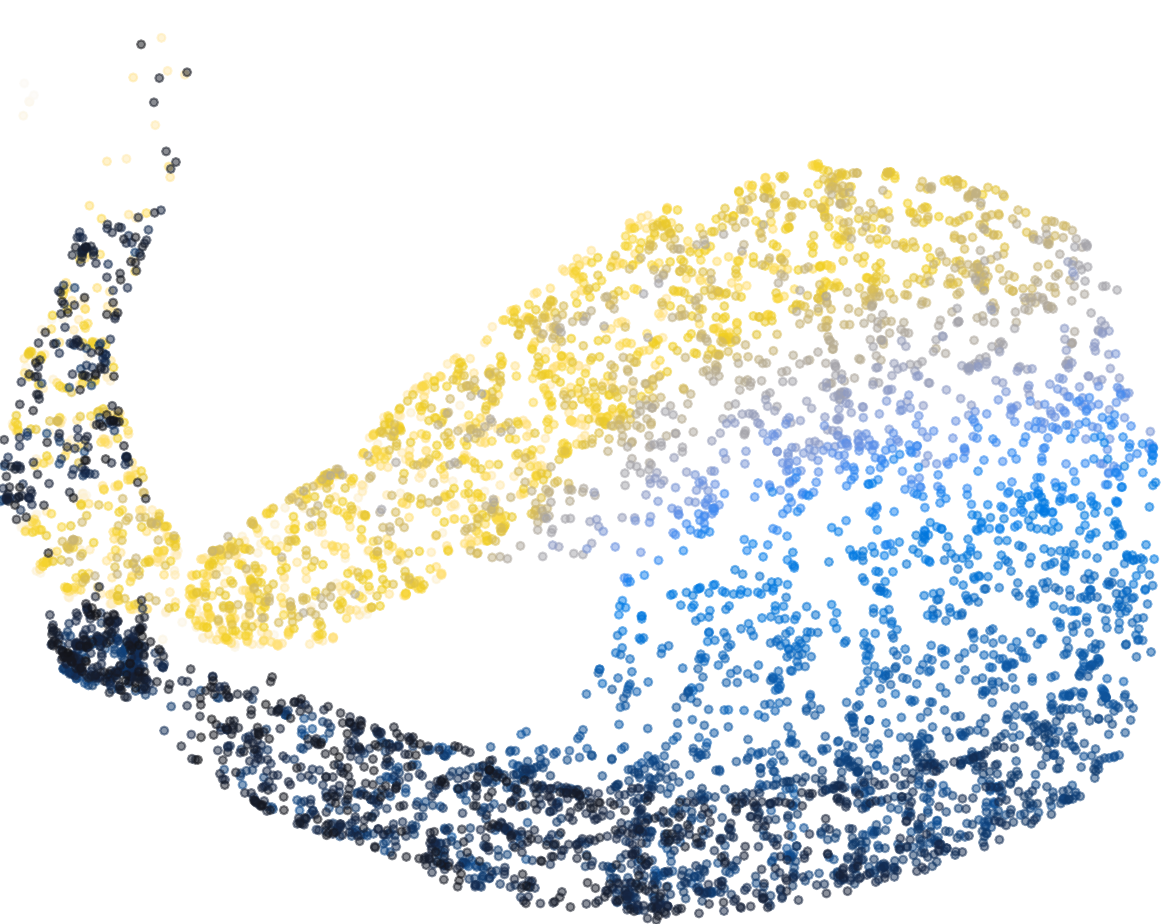}}%
    \hspace{.2cm}%
    \subfloat[Base station 8]{\includegraphics[width=0.15\linewidth]{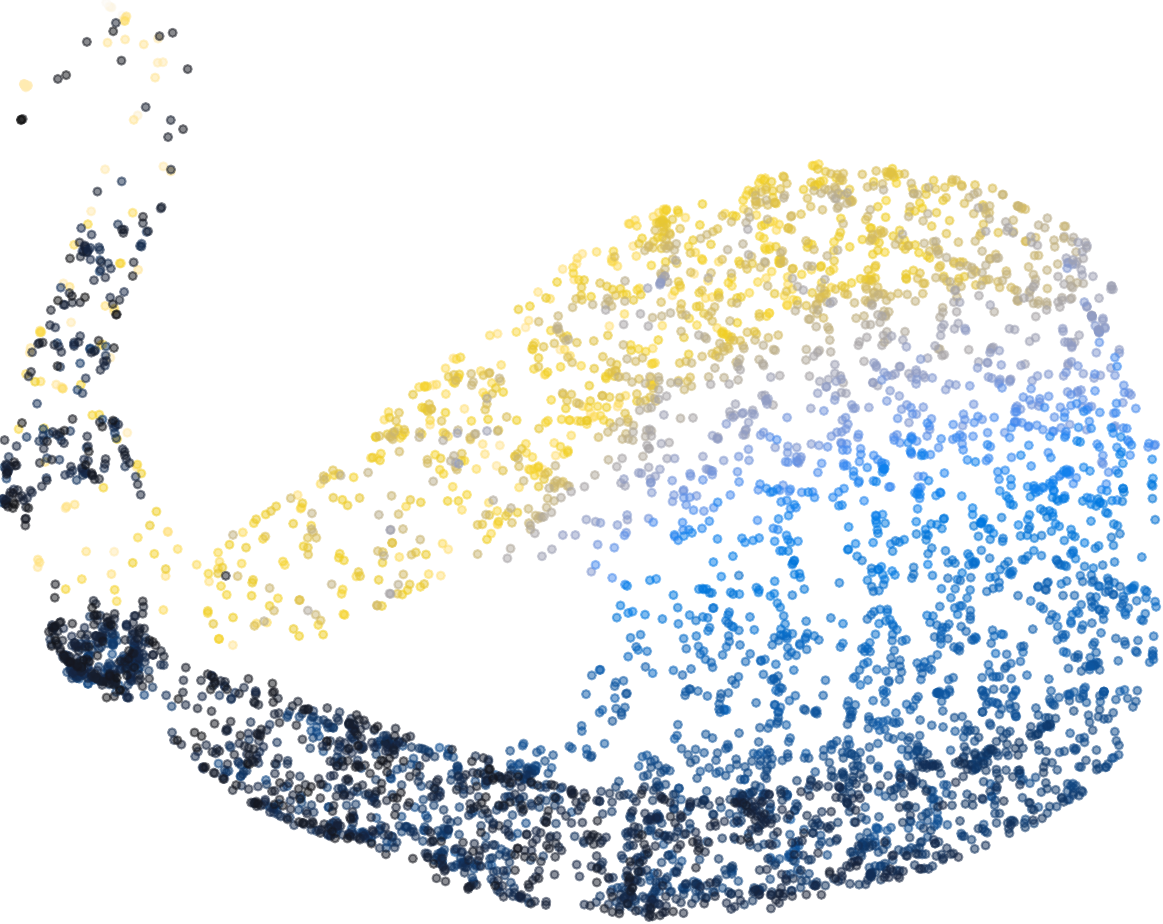}}%
    \hspace{.2cm}%
    \subfloat[Base station 6]{\includegraphics[width=0.15\linewidth]{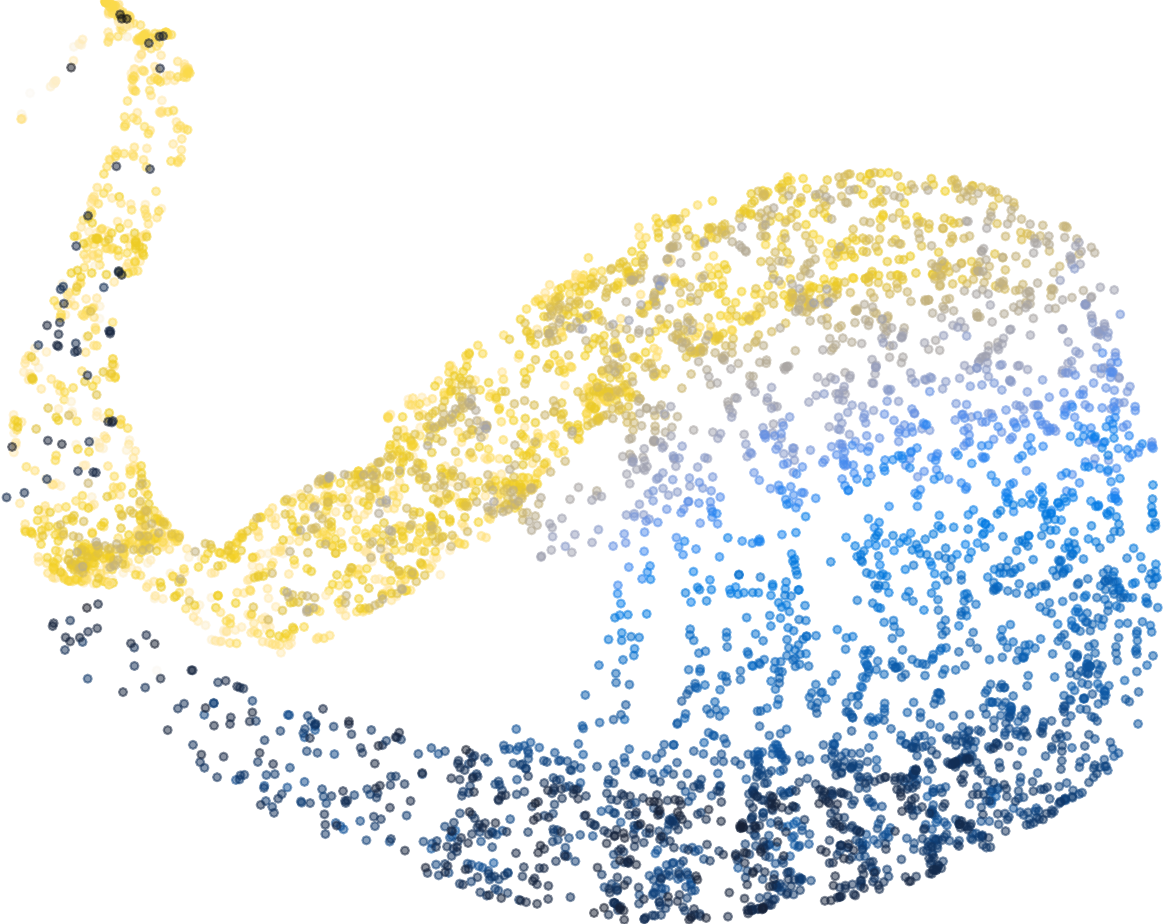}}%
    \hspace{.2cm}%
    \subfloat[Base station 4]{\includegraphics[width=0.15\linewidth]{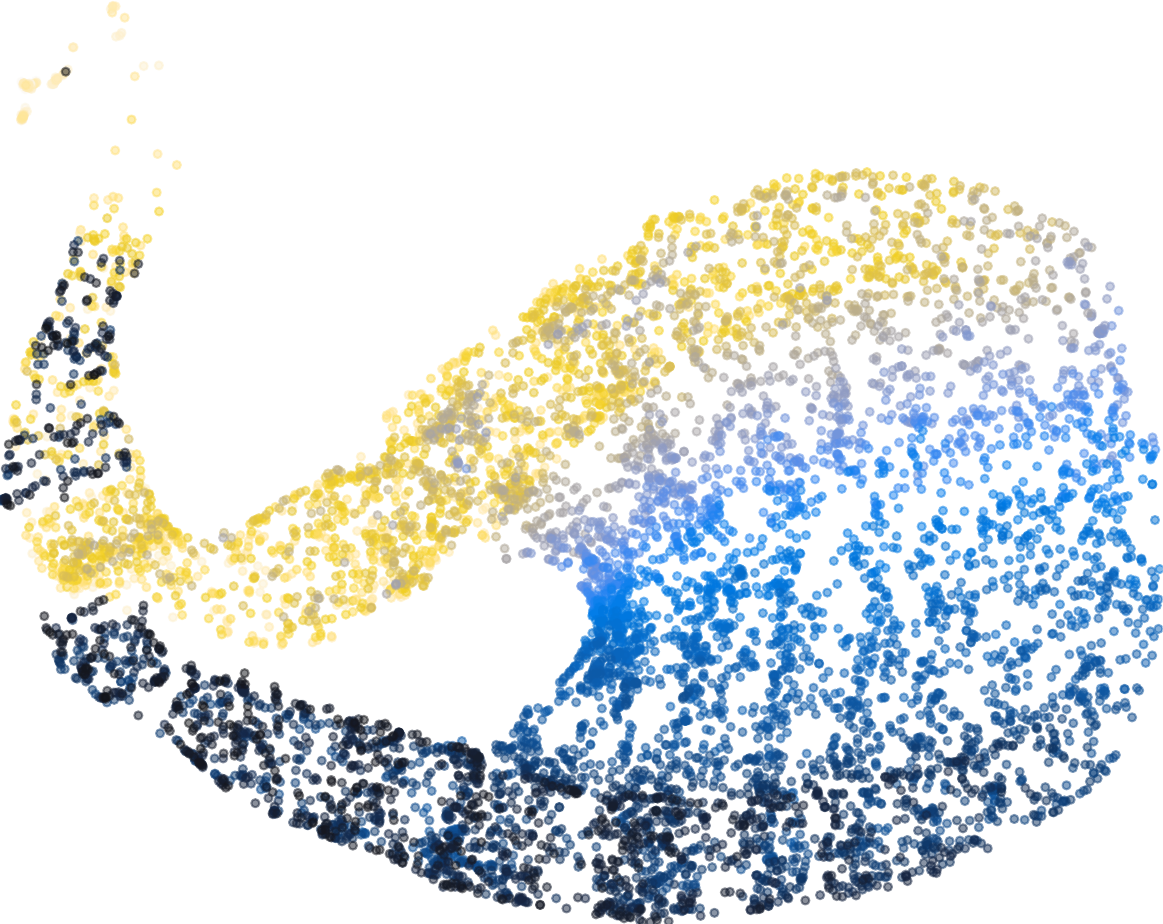}}%
    \hspace{.2cm}%
    \subfloat[Base station 2]{\includegraphics[width=0.15\linewidth]{figures/deepmimo_umap_bs_2.pdf}}%
    \hspace{.2cm}%
    \subfloat[Base station 11]{\includegraphics[width=0.15\linewidth]{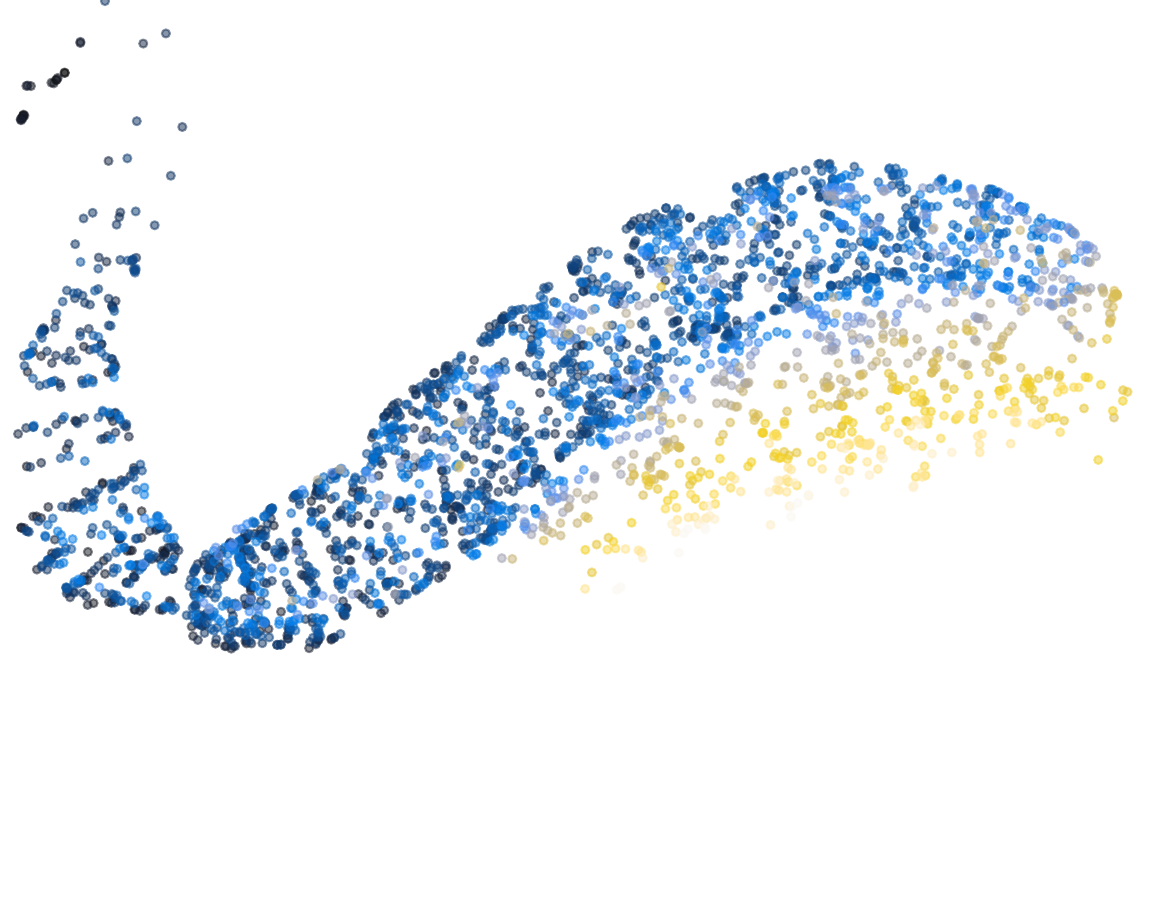}}%
    \hspace{.2cm}%
    \subfloat[Base station 9]{\includegraphics[width=0.15\linewidth]{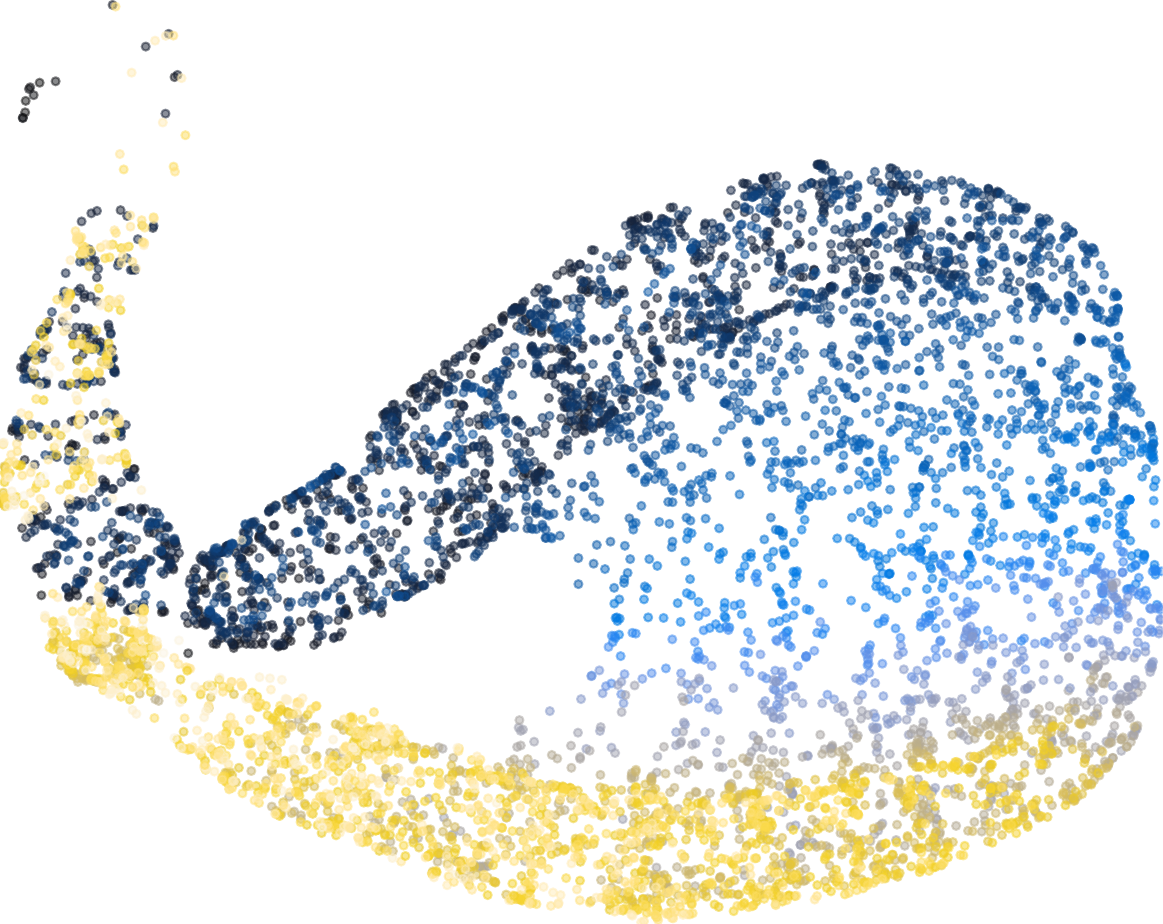}}%
    \hspace{.2cm}%
    \subfloat[Base station 7]{\includegraphics[width=0.15\linewidth]{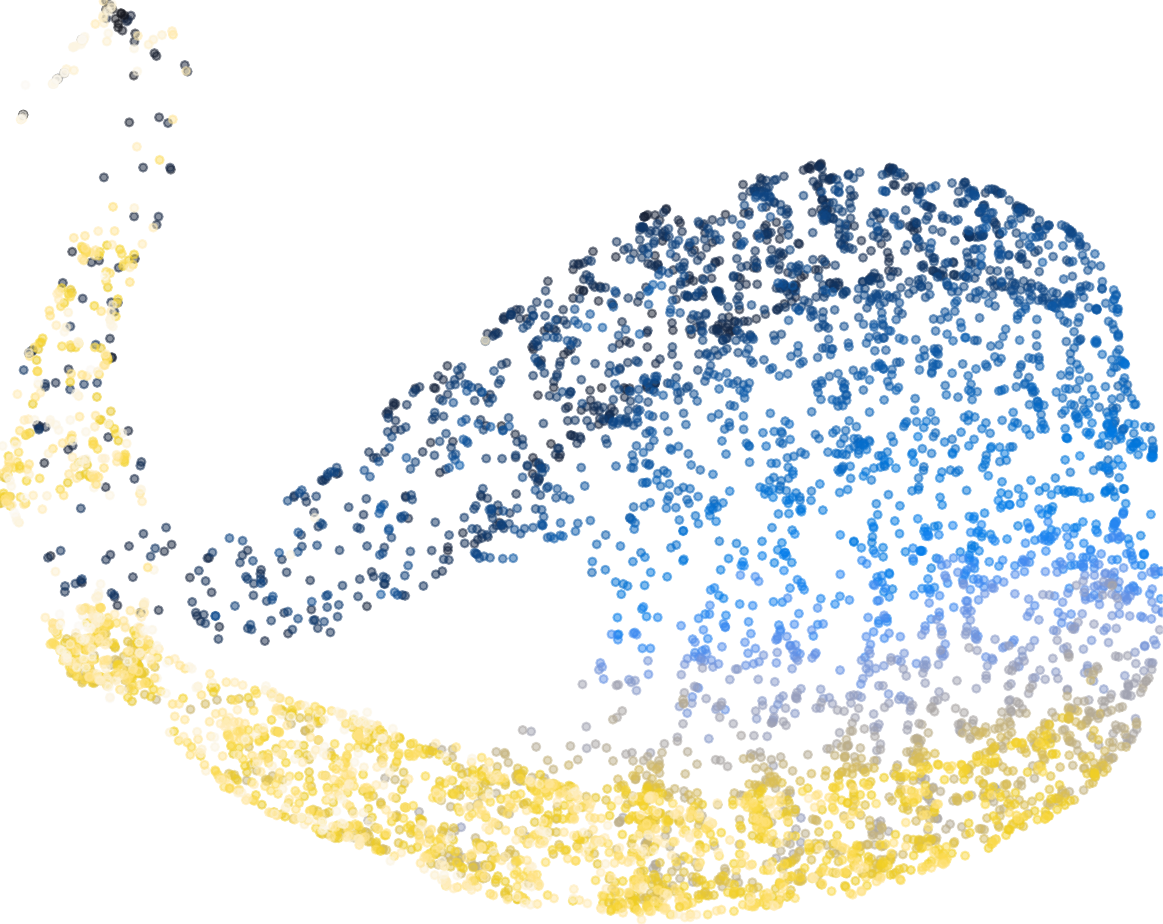}}%
    \hspace{.2cm}%
    \subfloat[Base station 5]{\includegraphics[width=0.15\linewidth]{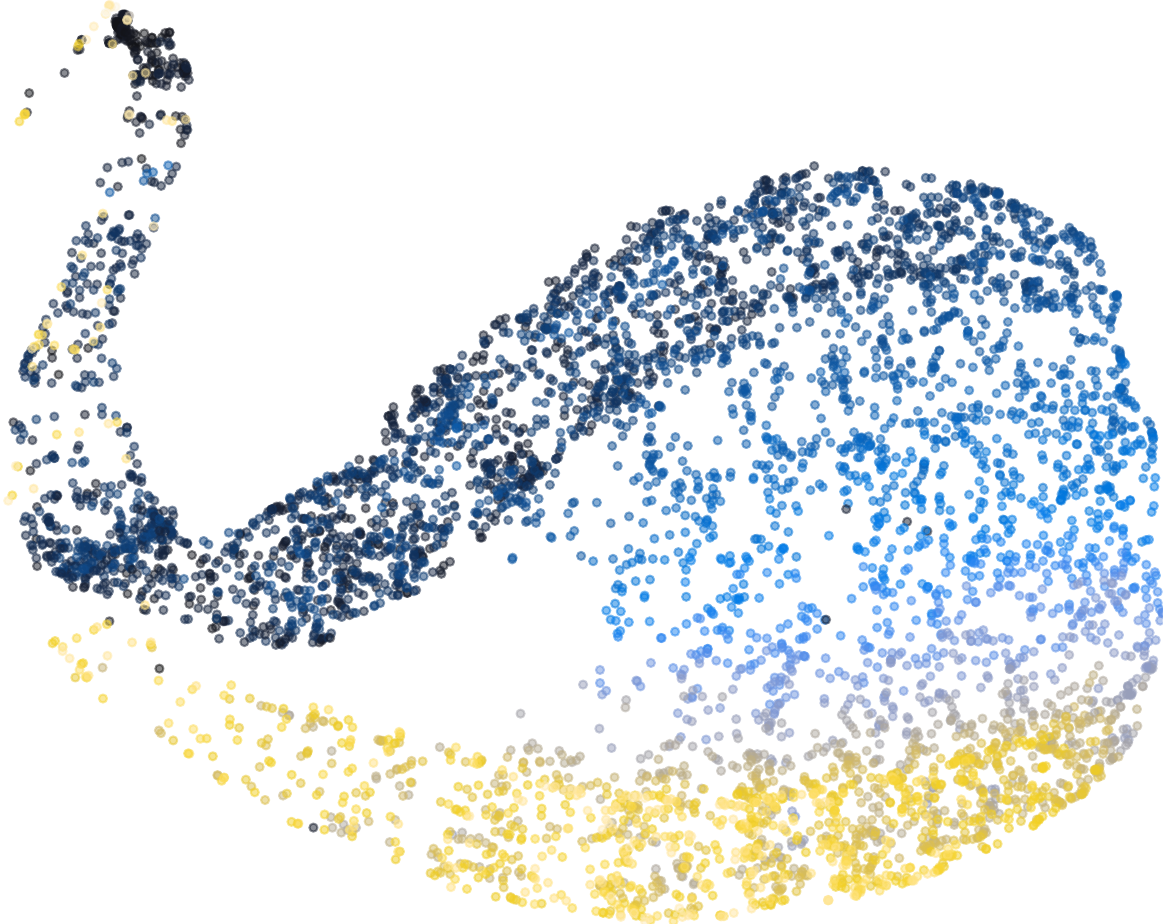}}%
    \hspace{.2cm}%
    \subfloat[Base station 3]{\includegraphics[width=0.15\linewidth]{figures/deepmimo_umap_bs_3.pdf}}%
    \hspace{.2cm}%
    \subfloat[Base station 1]{\includegraphics[width=0.15\linewidth]{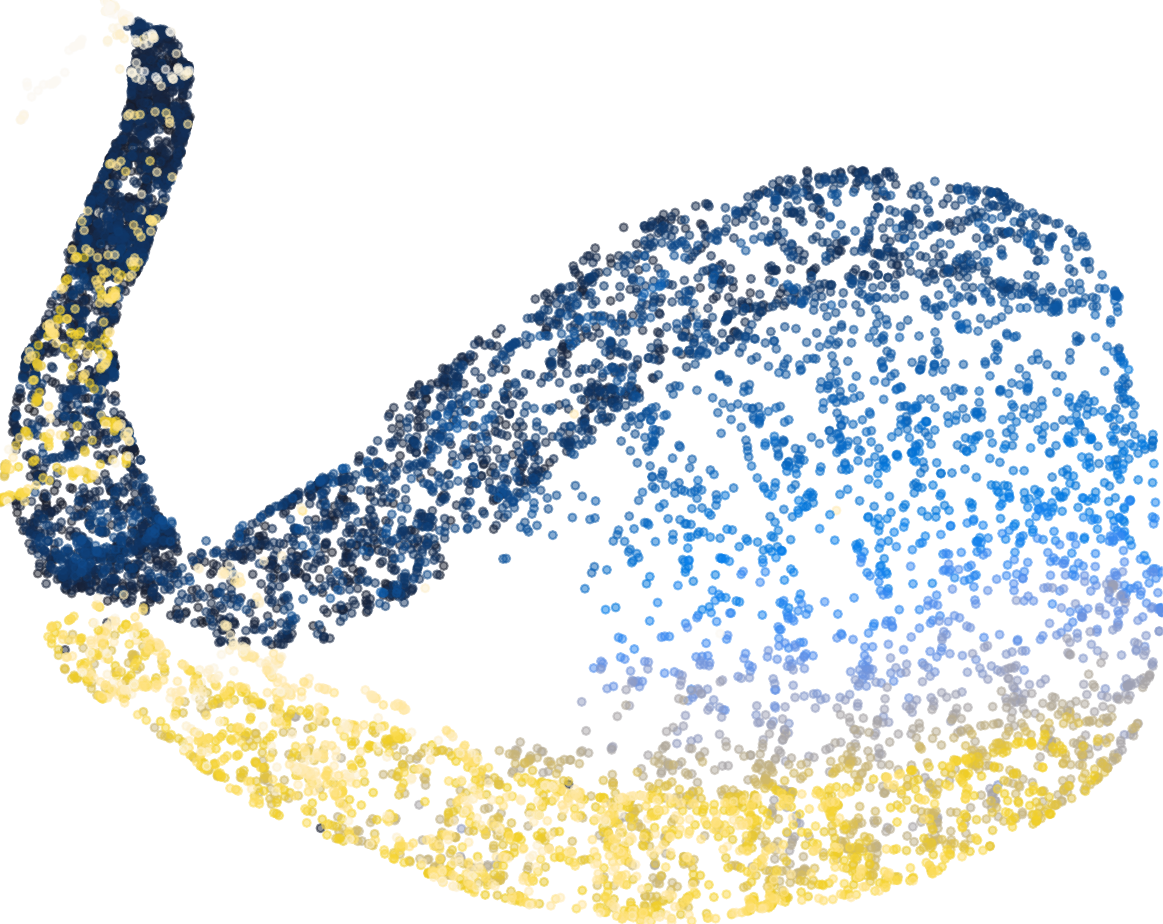}}%
    \caption{\textbf{Visualization of the \deepmimo{}~\cite{DBLP:journals/corr/abs-1902-06435,Remcom} dataset} using \gls{UMAP}~\cite{mcinnes2018umap} for all base stations. Color represents the target beamforming index. An overview map of the dataset is shown in~\cref{fig:map}.}
    \label{fig:umapbs3all}
\end{figure*}

\begin{table}[!ht]
    \centering
    \caption[\deepmimo{} results for \(\gls{principalcomponents} = 56\) and \(\gls{J} = 3\)]{\textbf{\deepmimo{} results} for \(\gls{principalcomponents} = 56\) and \(\gls{J} = 3\).}\label{tbl:deepmimo}
    \input{data/tables/tbl_deepmimo.tex}
\end{table}

\begin{table}[!ht]
    \centering
    \caption{\textbf{Welch's t-test~\cite{welch1947generalization} comparing between \pacfl{} and no data clustering algorithm with respect to accuracy for \deepmimo{}.} $p$-values adjusted with Benjamini--Hochberg \gls{FDR} correction~\cite{benjamini1995controlling} ($\alpha = 0.05$). Blue shading indicates improvement over baseline.}\label{tbl:deepmimo_pacfl_accuracy}
    \input{data/tables/tbl_pacfl_comp_deepmimo_accuracy.tex}
\end{table}

\begin{table}[!ht]
    \centering
    \caption{\textbf{Kolmogorov--Smirnov~\cite{massey1951kolmogorov} test comparing between \pacfl{} and no data clustering algorithm with respect to accuracy for \deepmimo{}.} $p$-values adjusted with Benjamini--Hochberg \gls{FDR} correction~\cite{benjamini1995controlling} ($\alpha = 0.05$).}\label{tbl:deepmimo_pacfl_accuracy_ks}
    \input{data/tables/tbl_pacfl_comp_deepmimo_accuracy_ks.tex}
\end{table}

\begin{table}[!ht]
    \centering
    \caption{\textbf{Welch's t-test~\cite{welch1947generalization} comparing between \ifca{} without data
    clustering algorithm to our methods with respect to accuracy for
    \deepmimo{}.} $p$-values adjusted with Benjamini--Hochberg \gls{FDR}
    correction~\cite{benjamini1995controlling} ($\alpha = 0.05$). Blue shading
    indicates improvement over baseline.}\label{tbl:deepmimo_ifca_accuracy}
    \input{data/tables/tbl_ifca_comp_deepmimo_accuracy.tex}
\end{table}

\begin{table}[!ht]
    \centering
    \caption{\textbf{Welch's t-test~\cite{welch1947generalization} comparing between \ifca{} without data
    clustering algorithm to our methods using \gls{PACFL} with respect to accuracy for
    \deepmimo{}.} $p$-values adjusted with Benjamini--Hochberg \gls{FDR}
    correction~\cite{benjamini1995controlling} ($\alpha = 0.05$). Blue shading
    indicates improvement over baseline.}\label{tbl:deepmimo_ifca_pacfl_accuracy}
    \input{data/tables/tbl_ifca_pacfl_comp_deepmimo_accuracy.tex}
\end{table}

\begin{table}[!ht]
    \centering
    \caption{\textbf{Welch's t-test~\cite{welch1947generalization} comparing between \pacfl{} and no data clustering algorithm with respect to communication rounds for \deepmimo{}.}
    $p$-values adjusted with Benjamini--Hochberg \gls{FDR} correction~\cite{benjamini1995controlling} ($\alpha = 0.05$). Blue shading indicates improvement over baseline.}\label{tbl:deepmimo_pacfl_rounds}
    \input{data/tables/tbl_pacfl_comp_deepmimo_rounds.tex}
\end{table}

\begin{table}[!ht]
    \centering
    \caption{\textbf{Kolmogorov--Smirnov~\cite{massey1951kolmogorov} test comparing between \pacfl{} and no data clustering algorithm with respect to communication rounds for \deepmimo{}.} $p$-values adjusted with Benjamini--Hochberg \gls{FDR} correction~\cite{benjamini1995controlling} ($\alpha = 0.05$).}\label{tbl:deepmimo_pacfl_rounds_ks}
    \input{data/tables/tbl_pacfl_comp_deepmimo_rounds_ks.tex}
\end{table}

\begin{table}[!ht]
    \centering
    \caption{\textbf{Welch's t-test~\cite{welch1947generalization} comparing \ifca{} without data clustering algorithm to our methods with respect to communication rounds for \deepmimo{}.}
    $p$-values adjusted with Benjamini--Hochberg \gls{FDR} correction~\cite{benjamini1995controlling} ($\alpha = 0.05$). Blue shading indicates improvement over baseline.}\label{tbl:deepmimo_ifca_rounds}
    \input{data/tables/tbl_ifca_comp_deepmimo_rounds.tex}
\end{table}

\begin{table}[!ht]
    \centering
    \caption{\textbf{Welch's t-test~\cite{welch1947generalization} comparing between \ifca{} without data
    clustering algorithm to our methods using \gls{PACFL} with respect to communication rounds for
    \deepmimo{}.} $p$-values adjusted with Benjamini--Hochberg \gls{FDR}
    correction~\cite{benjamini1995controlling} ($\alpha = 0.05$). Blue shading
    indicates improvement over baseline.}\label{tbl:deepmimo_ifca_pacfl_rounds}
    \input{data/tables/tbl_ifca_pacfl_comp_deepmimo_rounds.tex}
\end{table}

\begin{table}[!ht]
    \centering
    \caption{\textbf{Welch's t-test~\cite{welch1947generalization} comparing between \pacfl{} and no data clustering algorithm with respect to \gls{DL} data volume for \deepmimo{}.}
    $p$-values adjusted with Benjamini--Hochberg \gls{FDR} correction~\cite{benjamini1995controlling} ($\alpha = 0.05$). Blue shading indicates improvement over baseline.}\label{tbl:deepmimo_pacfl_totaldatadl}
    \input{data/tables/tbl_pacfl_comp_deepmimo_totaldatadl.tex}
\end{table}

\begin{table}[!ht]
    \centering
    \caption{\textbf{Welch's t-test~\cite{welch1947generalization} comparing between \ifca{} without data clustering algorithm to our methods with respect to \gls{DL} data volume for \deepmimo{}.} $p$-values adjusted with Benjamini--Hochberg \gls{FDR} correction~\cite{benjamini1995controlling} ($\alpha = 0.05$). Blue shading indicates improvement over baseline.}\label{tbl:deepmimo_ifca_totaldatadl}
    \input{data/tables/tbl_ifca_comp_deepmimo_totaldatadl.tex}
\end{table}

\begin{table}[!ht]
    \centering
    \caption{\textbf{Kolmogorov--Smirnov~\cite{massey1951kolmogorov} test comparing between \pacfl{} and no data clustering algorithm with respect to total \gls{DL} data volume for \deepmimo{}.} $p$-values adjusted with Benjamini--Hochberg \gls{FDR} correction~\cite{benjamini1995controlling} ($\alpha = 0.05$).}\label{tbl:deepmimo_pacfl_totaldatadl_ks}
    \input{data/tables/tbl_pacfl_comp_deepmimo_totaldatadl_ks.tex}
\end{table}

\begin{table}[!ht]
    \centering
    \caption{\textbf{Welch's t-test~\cite{welch1947generalization} comparing between \ifca{} without data
    clustering algorithm to our methods using \gls{PACFL} with respect to total \gls{DL} data volume for
    \deepmimo{}.} $p$-values adjusted with Benjamini--Hochberg \gls{FDR}
    correction~\cite{benjamini1995controlling} ($\alpha = 0.05$). Blue shading
    indicates improvement over baseline.}\label{tbl:deepmimo_ifca_pacfl_totaldatadl}
    \input{data/tables/tbl_ifca_pacfl_comp_deepmimo_totaldatadl.tex}
\end{table}

%% file: chapters/appendix_cifar10rot.tex
\clearpage\subsection{Detailed results for \rotatedcifar{}}
Here follows detailed results for the \rotatedcifar{} dataset, see~\cref{sec:datasets}.

\begin{table}[!ht]
    \centering
    \caption{\textbf{\rotatedcifar{} results} for \(\gls{principalcomponents} = 195\) and \(\gls{J} = 2\).}\label{tbl:cifar10rot}
    \input{data/tables/tbl_cifar10rot.tex}
\end{table}

\begin{table}[!ht]
    \centering
    \caption{\textbf{Welch's t-test~\cite{welch1947generalization} comparing between \pacfl{} and no data clustering algorithm with respect to accuracy for \rotatedcifar{}.} $p$-values adjusted with Benjamini--Hochberg \gls{FDR} correction~\cite{benjamini1995controlling} ($\alpha = 0.05$). Blue shading indicates improvement over baseline.}\label{tbl:cifar10rot_pacfl_accuracy}
    \input{data/tables/tbl_pacfl_comp_cifar10rot_accuracy.tex}
\end{table}

\begin{table}[!ht]
    \centering
    \caption{\textbf{Kolmogorov--Smirnov~\cite{massey1951kolmogorov} test comparing between \pacfl{} and no data clustering algorithm with respect to accuracy for \rotatedcifar{}.} $p$-values adjusted with Benjamini--Hochberg \gls{FDR} correction~\cite{benjamini1995controlling} ($\alpha = 0.05$).}\label{tbl:cifar10rot_pacfl_accuracy_ks}
    \input{data/tables/tbl_pacfl_comp_cifar10rot_accuracy_ks.tex}
\end{table}

\begin{table}[!ht]
    \centering
    \caption{\textbf{Welch's t-test~\cite{welch1947generalization} comparing between \ifca{} without data clustering algorithm to our methods with respect to accuracy for \rotatedcifar{}.} $p$-values adjusted with Benjamini--Hochberg \gls{FDR} correction~\cite{benjamini1995controlling} ($\alpha = 0.05$). Blue shading indicates improvement over baseline.}\label{tbl:cifar10rot_ifca_accuracy}
    \input{data/tables/tbl_ifca_comp_cifar10rot_accuracy.tex}
\end{table}

\begin{table}[!ht]
    \centering
    \caption{\textbf{Welch's t-test~\cite{welch1947generalization} comparing between \ifca{} without data
    clustering algorithm to our methods using \gls{PACFL} with respect to accuracy for
    \rotatedcifar{}.} $p$-values adjusted with Benjamini--Hochberg \gls{FDR}
    correction~\cite{benjamini1995controlling} ($\alpha = 0.05$). Blue shading
    indicates improvement over baseline.}\label{tbl:cifar10rot_ifca_pacfl_accuracy}
    \input{data/tables/tbl_ifca_pacfl_comp_cifar10rot_accuracy.tex}
\end{table}

\begin{table}[!ht]
    \centering
    \caption{\textbf{Welch's t-test~\cite{welch1947generalization} comparing between \pacfl{} and no data clustering algorithm with respect to communication rounds for \rotatedcifar{}.} $p$-values adjusted with Benjamini--Hochberg \gls{FDR} correction~\cite{benjamini1995controlling} ($\alpha = 0.05$). Blue shading indicates improvement over baseline.}\label{tbl:cifar10rot_pacfl_rounds}
    \input{data/tables/tbl_pacfl_comp_cifar10rot_rounds.tex}
\end{table}

\begin{table}[!ht]
    \centering
    \caption{\textbf{Kolmogorov--Smirnov~\cite{massey1951kolmogorov} test comparing between \pacfl{} and no data clustering algorithm with respect to communication rounds for \rotatedcifar{}.} $p$-values adjusted with Benjamini--Hochberg \gls{FDR} correction~\cite{benjamini1995controlling} ($\alpha = 0.05$).}\label{tbl:cifar10rot_pacfl_rounds_ks}
    \input{data/tables/tbl_pacfl_comp_cifar10rot_rounds_ks.tex}
\end{table}

\begin{table}[!ht]
    \centering
    \caption{\textbf{Welch's t-test~\cite{welch1947generalization} comparing between \ifca{} without data clustering algorithm to our methods with respect to communication rounds for \rotatedcifar{}.} $p$-values adjusted with Benjamini--Hochberg \gls{FDR} correction~\cite{benjamini1995controlling} ($\alpha = 0.05$). Blue shading indicates improvement over baseline.}\label{tbl:cifar10rot_ifca_rounds}
    \input{data/tables/tbl_ifca_comp_cifar10rot_rounds.tex}
\end{table}

\begin{table}[!ht]
    \centering
    \caption{\textbf{Welch's t-test~\cite{welch1947generalization} comparing between \ifca{} without data
    clustering algorithm to our methods using \gls{PACFL} with respect to communication rounds for
    \rotatedcifar{}.} $p$-values adjusted with Benjamini--Hochberg \gls{FDR}
    correction~\cite{benjamini1995controlling} ($\alpha = 0.05$). Blue shading
    indicates improvement over baseline.}\label{tbl:cifar10rot_ifca_pacfl_rounds}
    \input{data/tables/tbl_ifca_pacfl_comp_cifar10rot_rounds.tex}
\end{table}

\begin{table}[!ht]
    \centering
    \caption{\textbf{Welch's t-test~\cite{welch1947generalization} comparing between \pacfl{} and no data clustering algorithm with respect to \gls{DL} data volume for \rotatedcifar{}.}
    $p$-values adjusted with Benjamini--Hochberg \gls{FDR} correction~\cite{benjamini1995controlling} ($\alpha = 0.05$). Blue shading indicates improvement over baseline.}\label{tbl:cifar10rot_pacfl_totaldatadl}
    \input{data/tables/tbl_pacfl_comp_cifar10rot_totaldatadl.tex}
\end{table}

\begin{table}[!ht]
    \centering
    \caption{\textbf{Welch's t-test~\cite{welch1947generalization} comparing between \ifca{} without data clustering algorithm to our methods with respect to \gls{DL} data volume for \rotatedcifar{}.} $p$-values adjusted with Benjamini--Hochberg \gls{FDR} correction~\cite{benjamini1995controlling} ($\alpha = 0.05$). Blue shading indicates improvement over baseline.}\label{tbl:cifar10rot_ifca_totaldatadl}
    \input{data/tables/tbl_ifca_comp_cifar10rot_totaldatadl.tex}
\end{table}

\begin{table}[!ht]
    \centering
    \caption{\textbf{Kolmogorov--Smirnov~\cite{massey1951kolmogorov} test comparing between \pacfl{} and no data clustering algorithm with respect to total \gls{DL} data volume for \rotatedcifar{}.} $p$-values adjusted with Benjamini--Hochberg \gls{FDR} correction~\cite{benjamini1995controlling} ($\alpha = 0.05$).}\label{tbl:cifar10rot_pacfl_totaldatadl_ks}
    \input{data/tables/tbl_pacfl_comp_cifar10rot_totaldatadl_ks.tex}
\end{table}

\begin{table}[!ht]
    \centering
    \caption{\textbf{Welch's t-test~\cite{welch1947generalization} comparing between \ifca{} without data
    clustering algorithm to our methods using \gls{PACFL} with respect to total \gls{DL} data volume for
    \rotatedcifar{}.} $p$-values adjusted with Benjamini--Hochberg \gls{FDR}
    correction~\cite{benjamini1995controlling} ($\alpha = 0.05$). Blue shading
    indicates improvement over baseline.}\label{tbl:cifar10rot_ifca_pacfl_totaldatadl}
    \input{data/tables/tbl_ifca_pacfl_comp_cifar10rot_totaldatadl.tex}
\end{table}

%% file: chapters/appendix_femnist.tex
\clearpage\subsection{Detailed results for \femnist}
Here follows detailed results for the \femnist{} dataset, see~\cref{sec:datasets}.

\begin{table}[!ht]
    \centering
    \caption{\textbf{\femnist{} results} for \(\gls{principalcomponents} = 4\) and \(\gls{J} = 12\).}\label{tbl:femnist}
    \input{data/tables/tbl_femnist.tex}
\end{table}

\begin{table}[!ht]
    \centering
    \caption{\textbf{Welch's t-test~\cite{welch1947generalization} comparing between \pacfl{} and no data clustering algorithm with respect to accuracy for \femnist{}.} $p$-values adjusted with Benjamini--Hochberg \gls{FDR} correction~\cite{benjamini1995controlling} ($\alpha = 0.05$). Blue shading indicates improvement over baseline.}\label{tbl:femnist_pacfl_accuracy}
    \input{data/tables/tbl_pacfl_comp_femnist_accuracy.tex}
\end{table}

\begin{table}[!ht]
    \centering
    \caption{\textbf{Kolmogorov--Smirnov~\cite{massey1951kolmogorov} test comparing between \pacfl{} and no data clustering algorithm with respect to accuracy for \femnist{}.} $p$-values adjusted with Benjamini--Hochberg \gls{FDR} correction~\cite{benjamini1995controlling} ($\alpha = 0.05$).}\label{tbl:femnist_pacfl_accuracy_ks}
    \input{data/tables/tbl_pacfl_comp_femnist_accuracy_ks.tex}
\end{table}

\begin{table}[!ht]
    \centering
    \caption{\textbf{Welch's t-test~\cite{welch1947generalization} comparing between \ifca{} without data clustering algorithm to our methods with respect to accuracy for \femnist{}.} $p$-values adjusted with Benjamini--Hochberg \gls{FDR} correction~\cite{benjamini1995controlling} ($\alpha = 0.05$). Blue shading indicates improvement over baseline.}\label{tbl:femnist_ifca_accuracy}
    \input{data/tables/tbl_ifca_comp_femnist_accuracy.tex}
\end{table}

\begin{table}[!ht]
    \centering
    \caption{\textbf{Welch's t-test~\cite{welch1947generalization} comparing between \ifca{} without data
    clustering algorithm to our methods using \gls{PACFL} with respect to accuracy for
    \femnist{}.} $p$-values adjusted with Benjamini--Hochberg \gls{FDR}
    correction~\cite{benjamini1995controlling} ($\alpha = 0.05$). Blue shading
    indicates improvement over baseline.}\label{tbl:femnist_ifca_pacfl_accuracy}
    \input{data/tables/tbl_ifca_pacfl_comp_femnist_accuracy.tex}
\end{table}

\begin{table}[!ht]
    \centering
    \caption{\textbf{Welch's t-test~\cite{welch1947generalization} comparing between \pacfl{} and no data clustering algorithm with respect to communication rounds for \femnist{}.} $p$-values adjusted with Benjamini--Hochberg \gls{FDR} correction~\cite{benjamini1995controlling} ($\alpha = 0.05$). Blue shading indicates improvement over baseline.}\label{tbl:femnist_pacfl_rounds}
    \input{data/tables/tbl_pacfl_comp_femnist_rounds.tex}
\end{table}

\begin{table}[!ht]
    \centering
    \caption{\textbf{Kolmogorov--Smirnov~\cite{massey1951kolmogorov} test comparing between \pacfl{} and no data clustering algorithm with respect to communication rounds for \femnist{}.} $p$-values adjusted with Benjamini--Hochberg \gls{FDR} correction~\cite{benjamini1995controlling} ($\alpha = 0.05$).}\label{tbl:femnist_pacfl_rounds_ks}
    \input{data/tables/tbl_pacfl_comp_femnist_rounds_ks.tex}
\end{table}

\begin{table}[!ht]
    \centering
    \caption{\textbf{Welch's t-test~\cite{welch1947generalization} comparing between \ifca{} without data clustering algorithm to our methods with respect to communication rounds for \femnist{}.} $p$-values adjusted with Benjamini--Hochberg \gls{FDR} correction~\cite{benjamini1995controlling} ($\alpha = 0.05$). Blue shading indicates improvement over baseline.}\label{tbl:femnist_ifca_rounds}
    \input{data/tables/tbl_ifca_comp_femnist_rounds.tex}
\end{table}

\begin{table}[!ht]
    \centering
    \caption{\textbf{Welch's t-test~\cite{welch1947generalization} comparing between \ifca{} without data
    clustering algorithm to our methods using \gls{PACFL} with respect to communication rounds for
    \femnist{}.} $p$-values adjusted with Benjamini--Hochberg \gls{FDR}
    correction~\cite{benjamini1995controlling} ($\alpha = 0.05$). Blue shading
    indicates improvement over baseline.}\label{tbl:femnist_ifca_pacfl_rounds}
    \input{data/tables/tbl_ifca_pacfl_comp_femnist_rounds.tex}
\end{table}

\begin{table}[!ht]
    \centering
    \caption{\textbf{Welch's t-test~\cite{welch1947generalization} comparing between \pacfl{} and no data
    clustering algorithm with respect to \gls{DL} data volume for \femnist{}.}
    $p$-values adjusted with Benjamini--Hochberg \gls{FDR}
    correction~\cite{benjamini1995controlling} ($\alpha = 0.05$). Blue shading
    indicates improvement over baseline.}\label{tbl:femnist_pacfl_totaldatadl}
    \input{data/tables/tbl_pacfl_comp_femnist_totaldatadl.tex}
\end{table}

\begin{table}[!ht]
    \centering
    \caption{\textbf{Welch's t-test~\cite{welch1947generalization} comparing between \ifca{} without data
    clustering algorithm to our methods with respect to \gls{DL} data volume
    for \femnist{}.} $p$-values adjusted with Benjamini--Hochberg \gls{FDR}
    correction~\cite{benjamini1995controlling} ($\alpha = 0.05$). Blue shading
    indicates improvement over baseline.}\label{tbl:femnist_ifca_totaldatadl}
    \input{data/tables/tbl_ifca_comp_femnist_totaldatadl.tex}
\end{table}

\begin{table}[!ht]
    \centering
    \caption{\textbf{Kolmogorov--Smirnov~\cite{massey1951kolmogorov} test
    comparing between \pacfl{} and no data clustering algorithm with respect to
    total \gls{DL} data volume for \femnist{}.} $p$-values adjusted with
    Benjamini--Hochberg \gls{FDR} correction~\cite{benjamini1995controlling}
    ($\alpha = 0.05$).}\label{tbl:femnist_pacfl_totaldatadl_ks}
    \input{data/tables/tbl_pacfl_comp_femnist_totaldatadl_ks.tex}
\end{table}

\begin{table}[!ht]
    \centering
    \caption{\textbf{Welch's t-test~\cite{welch1947generalization} comparing between \ifca{} without data
    clustering algorithm to our methods using \gls{PACFL} with respect to total \gls{DL} data volume for
    \femnist{}.} $p$-values adjusted with Benjamini--Hochberg \gls{FDR}
    correction~\cite{benjamini1995controlling} ($\alpha = 0.05$). Blue shading
    indicates improvement over baseline.}\label{tbl:femnist_ifca_pacfl_totaldatadl}
    \input{data/tables/tbl_ifca_pacfl_comp_femnist_totaldatadl.tex}
\end{table}

%% file: chapters/hyperparams.tex
\section{Hyperparameters}\label{sec:hyperparams}

\begin{table}[!ht]
    \centering
    \caption[Hyperparameters for \femnist{}]{\textbf{Hyperparameters for \femnist{}}~\cite{caldas2018leaf,Cohen2017}.}
    \input{data/tables/tbl_hyperparams_femnist.tex}
\end{table}

\begin{table}[!ht]
    \centering
    \caption[Hyperparameters for \deepmimo{}]{\textbf{Hyperparameters for \deepmimo{}}~\cite{DBLP:journals/corr/abs-1902-06435,Remcom}.}
    \input{data/tables/tbl_hyperparams_deepmimo.tex}
\end{table}

\begin{table}[!ht]
    \centering
    \caption[Hyperparameters for \rotatedcifar{}]{\textbf{Hyperparameters for \rotatedcifar{}}~\cite{Krizhevsky2009}.}
    \input{data/tables/tbl_hyperparams_cifar10rot.tex}
\end{table}

\begin{table}[!ht]
    \centering
    \caption[Hyperparameters for \deepmimo{}]{\textbf{Hyperparameters for \deepmimo{}}~\cite{DBLP:journals/corr/abs-1902-06435,Remcom}.}
    \input{data/tables/tbl_hyperparams_deepmimo_sweep.tex}
\end{table}